\makeatletter
\def\input@path{{style/}}

\documentclass{vldb}
 \pdfoutput=1
\usepackage[ruled, lined, linesnumbered]{algorithm2e}
\SetAlFnt{\small}
\SetAlCapFnt{\small}

\usepackage[font=small, skip=4pt, bf]{caption}
\usepackage{subfigure}
\usepackage{algpseudocode}
\usepackage{graphicx}
\usepackage{balance}
\usepackage{color}
\usepackage{amssymb}
\usepackage{pifont}
\usepackage{mathabx}
\usepackage{bigstrut}
\usepackage{array}
\usepackage{booktabs}
\usepackage{soul}
\usepackage{hyperref}
\usepackage{float}
\usepackage{array}
\usepackage{afterpage}
\usepackage{xcolor}
\newcommand{\nop}[1]{}

\graphicspath{{images/}}
\newcommand{\Chi}[1]{\textcolor{red}{#1}}

\newcolumntype{L}[1]{>{\raggedright\let\newline\\\arraybackslash\hspace{0pt}}m{#1}}
\begin{document}

\newcommand{\reminder}[1]{\textbf{[** #1 **]}}  


\newcommand{\todo}[1]{\vspace{5 mm}\par \noindent
 \marginpar
{\textsc{ToDo}}
 \framebox{\begin{minipage}[c]{0.95 \textwidth}
 \tt #1 \end{minipage}}\vspace{5 mm}\par}

\newcommand{\eat}[1]{}

\def\papernumber #1 raised #2 {
  \vspace{-#2}
  \vbox to 0pt{\framebox{\bf Paper Number: #1}}
}


\newcommand{\group}{g}
\newcommand{\groupw}{\group^w}
\newcommand{\minsup}{minsup}
\newcommand{\UniRank}{UniRank}
\newcommand{\ST}{PTWCorr}
\newcommand{\UniWCorr}{UniWCorr}
\newcommand{\DNG}{DeTeNG}
\newcommand{\UniCorr}{UniCorr}
\newcommand{\PODD}{GMoDD}
\newcommand{\gsim}{\Omega}
\newcommand{\tsize}{T}
\newcommand{\cluster}{\mathcal{C}}
\newcommand{\pair}{\mathcal{S}}
\newcommand{\SEPR}{\hspace*{0.6cm}}
\newcommand{\sepr}{\hspace*{0.4cm}}
\newcommand{\smsep}{\hspace*{0.2cm}}
\newcommand{\s}{\hspace*{2mm}}
\newcommand{\noparag}{\hspace*{-\parindent}}
\newcommand{\parag}{\hspace*{ \parindent}}
\newcommand{\smgap}{-5pt}
\newcommand{\para}[1]{\noindent\textbf{#1}}

\newcommand{\un}{\underline}
\newcommand{\pf}{\sf}                   
\newcommand{\fs}{\small}                
\newcommand{\schf}{\footnotesize\tt}    
\newcommand{\PFB}[1]{\mbox{\pf \footnotesize #1}}
\newcommand{\PFBs}[1]{\mbox{\pf \scriptsize #1}}
\renewcommand{\ttdefault}{pcr}

\newcommand{\SetOf}{\PFB{{SetOf}}}
\newcommand{\Rcd}{\PFB{{Rcd}}}
\newcommand{\Choice}{\PFB{{Choice}}}
\newcommand{\Str}{\PFB{{str}}}
\newcommand{\Int}{\PFB{{int}}}
\newcommand{\Flt}{\PFB{{float}}}

\newcommand{\Set}{\PFB{\un{set}}}
\newcommand{\Let}{\PFB{\un{let}}}
\newcommand{\mf}{\PFB{\un{MF}}}
\newcommand{\mr}{\PFB{\un{MR}}}
\newcommand{\In}{~\PFB{\un{in}}~}
\newcommand{\Where}{\PFB{\un{where}}}
\newcommand{\With}{\PFB{\un{with}}}
\newcommand{\Select}{\PFB{\un{select}}}
\newcommand{\Orderby}{\PFB{\un{order by}}}
\newcommand{\Groupby}{\PFB{\un{group by}}}
\newcommand{\Limit}{\PFB{\un{limit}}}
\newcommand{\From}{\PFB{\un{from}}}
\newcommand{\EAnd}{~\PFB{\un{and}}~}
\newcommand{\Tuple}{\PFB{~\un{struct}~}}
\newcommand{\Exists}{\PFB{\un{exists}}}
\newcommand{\Element}{\PFB{~\un{element}~}}
\newcommand{\Every}{\PFB{\un{every}}}
\newcommand{\Satisfies}{\PFB{\un{satisfies}}}

\newcommand{\Ins}{~\PFBs{\un{in}}~}
\newcommand{\Wheres}{\PFBs{\un{where}}}
\newcommand{\Withs}{\PFBs{\un{with}}}
\newcommand{\Selects}{\PFBs{\un{select}}}
\newcommand{\Fors}{\PFBs{\un{for}}}
\newcommand{\Returns}{\PFBs{\un{return}}}
\newcommand{\Froms}{\PFBs{\un{from}}}
\newcommand{\EAnds}{~\PFBs{\un{and}}~}
\newcommand{\Tuples}{\PFBs{~\un{struct}~}}
\newcommand{\Existss}{\PFBs{\un{exists}}}
\newcommand{\Elements}{\PFBs{~\un{element}~}}
\newcommand{\Foreachs}{\PFBs{\un{foreach}}}
\newcommand{\Everys}{\PFBs{\un{every}}}
\newcommand{\Ifs}{\PFBs{\un{if}}}
\newcommand{\Thens}{\PFBs{\un{then}}}
\newcommand{\Satisfiess}{\PFBs{\un{satisfies}}}
\newcommand{\expl}{\mbox {\tt Expl}}
\newcommand{\itemsim}{\mbox {\tt ItemSim}}
\newcommand{\usersim}{\mbox {\tt UserSim}}
\newcommand{\calu}{\mbox {$\cal U$}}
\newcommand{\cali}{\mbox {$\cal I$}}
\newcommand{\nw}{\mbox {\tt Network}}
\newcommand{\ddu}{\mbox {$DD_u$}}
\newcommand{\ddju}{\mbox {$DD^J_u$}}
\newcommand{\ddcu}{\mbox {$DD^C_u$}}
\newcommand{\socrel}{\mbox {\tt SocRel}}

\newcommand{\from}[2]{{\sc from} {\bf #1: #2}}

\newcommand{\ch}{$\rightarrow$}
\newcommand{\mt}[1]{$#1$}
\newcommand{\mc}[1]{$\mathcal{#1}$}
\newcommand{\mcs}[2]{$\mathcal{#1}_{#2}$}

\newtheorem{definition}{Definition}

\newtheorem{Ob}{Observation}

\newenvironment{query}{
\footnotesize
\vspace*{\smgap}
\begin{tabbing} }
{\end{tabbing}
\normalsize\small\normalsize
\vspace*{\smgap}
}

\newenvironment{querys}{
\scriptsize
\vspace*{\smgap}
\begin{tabbing} }
{\end{tabbing}
\normalsize\small\normalsize
\vspace*{\smgap}
}

\newcommand{\secref}[1]{Section~\ref{#1}} 
\newcommand{\figref}[1]{Figure~\ref{#1}} 
\newcommand{\tbref}[1]{Table~\ref{#1}} 
\newcommand{\cass}{Hypothesis~}
\newcommand{\cequ}{Equation~}
\newcommand{\hide}[1]{} 
\newcommand{\ie}{{\sl i.e.}}
\newcommand{\eg}{{\sl e.g.}}
\newcommand{\etc}{{\sl etc.}}
\newcommand{\etal}{{\sl et al.}}
\newcommand{\adhoc}{{\sl ad hoc}}
\newtheorem{property}{Property}
\newtheorem{example}{Example}

\title{Scalable Topical Phrase Mining from Text Corpora}

%
%
%
%
%

\numberofauthors{1} 
%
\author{
%
%
\alignauthor
Ahmed El-Kishky$^{\dag}$, Yanglei Song$^{\dag}$, Chi Wang$^{\S}$, Clare R. Voss$^{\ddag}$, Jiawei Han$^{\dag}$\\
       \affaddr{$^{\dag}$Department of Computer Science The University of Illinois at Urbana Champaign}\\
       \affaddr{$^{\S}$Microsoft Research}\\
       \affaddr{$^{\ddag}$Computational \& Information Sciences Directorate, Army Research Laboratory}\\
       \affaddr{$^{\dag}$Urbana, IL, USA}, 
       \affaddr{$^{\S}$Redmond, WA, USA},
       \affaddr{$^{\ddag}$Adelphi, MD, USA}\\
       \email{\sf \{elkishk2, ysong44, hanj\}@illinois.edu, chiw@microsoft.com, clare.r.voss.civ@mail.mil}
}

\maketitle
\begin{abstract}
While most topic modeling algorithms model text corpora with unigrams, human interpretation often relies on inherent grouping of terms into phrases.  As such, we consider the problem of discovering topical phrases of mixed lengths. Existing work either performs post processing to the results of unigram-based topic models, or utilizes complex n-gram-discovery topic models. These methods generally produce low-quality topical phrases or suffer from poor scalability on even moderately-sized datasets. We propose a different approach that is both computationally efficient and effective. Our solution combines a novel phrase mining framework to segment a document into single and multi-word phrases, and a new topic model that operates on the induced document partition.   Our approach discovers high quality topical phrases with negligible extra cost to the bag-of-words topic model in a variety of datasets including research publication titles, abstracts, reviews, and news articles.

\end{abstract}
%




\section{Introduction}
\label{sec:intro}
In recent years, topic modeling has become a popular method for discovering the abstract `topics' that underly a collection of documents. A topic is typically modeled as a multinomial distribution over terms, and frequent terms related by a common theme are expected to have a large probability in a topic multinomial.
\begin{table}
\centering
\begin{tabular}{ @{} *2l @{} }
  \hline
 \emph{Terms} & \emph{Phrases} \\
  \hline
  search & information retrieval \\
  web & social networks \\
  retrieval & web search \\
  information & search engine \\
  based & support vector machine \\
  model & information extraction \\
  document & web page \\
  query & question answering \\
  text & text classification \\
  social & collaborative filtering \\
  user & topic model \\
  \hline
\end{tabular}
\caption{ Visualization of the topic of Information Retrieval, automatically constructed by ToPMine from titles of computer science papers published in DBLP (20Conf dataset).}
\label{tb:ir}
\end{table}
When latent topic multinomials are inferred, it is of interest to visualize these topics in order to facilitate human interpretation and exploration of the large amounts of unorganized text often found within text corpora. In addition, visualization provides a qualitative method of validating the inferred topic model~\cite{Chang:Boyd-Graber:Wang:Gerrish:Blei-2009}. A list of most probable unigrams is often used to describe individual topics, yet these unigrams often provide a hard-to-interpret or ambiguous representation of the topic. Augmenting unigrams with a list of probable phrases provides a more intuitively understandable and accurate description of a topic. This can be seen in the term/phrase visualization of an information retrieval topic in Table~\ref{tb:ir}. 

While topic models have clear application in facilitating understanding, organization, and exploration in large text collections such as those found in full-text databases, difficulty in interpretation and scalability issues have hindered adoption. Several attempts have been made to address the prevalent deficiency in visualizing topics using unigrams. These methods generally attempt to infer phrases and topics simultaneously by creating complex generative mechanism. The resultant models can directly output phrases and their latent topic assignment. Two such methods are Topical N-Gram and PD-LDA~\cite{wang2007topical, lindsey2012phrase}. While it is appealing to incorporate the phrase-finding element in the topical clustering process, these methods often suffer from high-complexity, and overall demonstrate poor scalability outside small datasets. 

Some other methods apply a post-processing step to unigram-based topic models~\cite{blei2009visualizing,danilevsky2014kert}. These methods assume that all words in a phrase will be assigned to a common topic, which, however, is not guaranteed by the topic model.

We propose a new methodology ToPMine that demonstrates both scalability compared to other topical phrase methods and interpretability. Because language exhibits the principle of non-compositionality, where a phrase's meaning is not derivable from its constituent words, under the `bag-of-words' assumption, phrases are decomposed, and a phrase's meaning may be lost~\cite{schone2001knowledge}. Our insight is that phrases need to be systematically assigned to topics. This insight motivates our partitioning of a document into phrases, then using these phrases as constraints to ensure all words are systematically placed in the same topic.

We perform topic modeling on phrases by  \emph{first mining phrases, segmenting each document into single and multi-word phrases, and then using the constraints from segmentation in our topic modeling}. First, to address the scalability issue, we develop an efficient phrase mining technique to extract frequent significant phrases and segment the text simultaneously. It uses frequent phrase mining and a statistical significance measure to segment the text while simultaneously filtering out false candidates phrases. Second, to ensure a systematic method of assigning latent topics to phrases, we propose a simple but effective topic model. By restricting all constituent terms within a phrase to share the same latent topic, we can assign a phrase the topic of its constituent words.

\begin{example}
\label{eg:seg}
By frequent phrase mining and context-specific statistical significance ranking, the following titles can be segmented as follows: 

\textbf{Title 1.} \textbf{[}Mining frequent patterns\textbf{]} without candidate generation: a \textbf{[}frequent pattern\textbf{]} tree approach.

\textbf{Title 2.} \textbf{[}Frequent pattern mining\textbf{]} : current status and future directions.

The tokens grouped together by \textbf{[]} are constrained to share the same topic assignment.

\end{example}
Our TopMine method has the following advantages.
\begin{itemize}
\parskip -0.2ex
\item Our phrase mining algorithm efficiently extracts candidate phrases and the necessary aggregate statistics needed to prune these candidate phrases. Requiring no domain knowledge or specific linguistic rulesets, our method is purely data-driven.
 
 \item Our method allows for an efficient and accurate filtering of false-candidate phrases. In title 1 of Example~\ref{eg:seg}, after merging `frequent' and `pattern', we only need to test whether `frequent pattern tree' is a \emph{significant} phrase in order to determine whether to keep `frequent pattern' as a phrase in this title. 
  \item Segmentation induces a `bag-of-phrases' representation for documents. We incorporate this as a constraint into our topic model eliminating the need for additional latent variables to find the phrases. The model complexity is reduced and the conformity of topic assignments within each phrase is maintained.
  
\end{itemize}

We state and analyze the problem in Section~\ref{sec:probDefinition}, followed by our proposed solution in Section~\ref{sec:method}. We present the key components of our solution, phrase mining and phrase-constrained topic modeling in Sections~\ref{sec:phrase} and \ref{sec:model}. In Section~\ref{sec:related}, we review the related work. Then we evaluate the proposed solution in Section~\ref{sec:exp}, and conclude in Section~\ref{sec:conclusion}.

\section{Problem Definition}
\label{sec:probDefinition}
The input is a corpus of $D$ documents, where $d$-th document is  a sequence of $\mathcal{N}_d$ tokens: $w_{d,i}, i=1,\dots,\mathcal{N}_d$. Let $N=\sum_{d=1}^D \mathcal{N}_d$. For convenience we index all the unique words in this corpus using a vocabulary of $V$ words. And $w_{d,i}=x, x\in\{1,\dots,V\}$ means that the $i$-th token in $d$-th document is the $x$-th word in the vocabulary. Throughout this paper we use `word $x$' to refer to the $x$-th word in the vocabulary.

Given a corpus and a number of topics as a parameter, our goal is to infer the corpus' underlying topics and visualize these topics in a human-interpretable representation using topical phrases. Statistically, a topic $k$ is characterized by a probability distribution $\phi_k$ over words. $\phi_{k,x}=p(x|k)\in [0,1]$ is the probability of seeing the word $x$ in topic $k$, and $\sum_{x=1}^V \phi_{k,x}=1$. For example, in a topic about the database research area, the probability of seeing ``database", ``system" and ``query" is high, and the probability of seeing ``speech", ``handwriting" and ``animation" is low. This characterization is advantageous in statistical modeling of text, but is weak in human interpretability. Unigrams may be ambiguous, especially across specific topics. For example, the word ``model" can mean different things depending on a topic - a model could be a member of the fashion industry or perhaps be part of a phrase such as ``topic model". Using of phrases helps avoid this ambiguity. 

\begin{definition} We formally define phrases and other necessary notation and terminology as follows: 
\begin{itemize}
	\item A \emph{phrase} is a sequence of contiguous tokens: \\$P$=$\{w_{d,i},...w_{d,i+n}\}\; n> 0$
	\item A \emph{partition} over $d$-th document is a sequence of phrases: $(P_{d,1}, \dots, P_{d,G_d} )\;G_d\geq 1$ s.t. the concatenation of the phrase instances is the original document.
\end{itemize}
\end{definition}

In Example~\ref{eg:seg}, we can see the importance of word proximity in phrase recognition. As such, we place a contiguity restriction on our phrases. \eat{Under this constraint, we have observed that phrases demonstrate strong interpretability, even under different word orders,\eat{. This can be easily seen in} for example the phrases \emph{`frequent pattern mining'} and \emph{`mining frequent patterns'} found in titles 1 and 2. Despite different word order, the phrases above relay the same meaning. This observation motivates our phrase representation as a multiset over words, which allows for the flexibility to model phrases such as \emph{`really really tired'} where a word can repeat. }

To illustrate an induced partition upon a text segment, we can note how the concatenation of all single and multi-word phrases in Title 1 will yield an ordered sequence of tokens representing the original title.

\subsection{Desired Properties}
We outline the desired properties of a topical phrase mining algorithm as follows:
\begin{itemize}
\item The lists of phrases demonstrate a coherent topic.
\item The phrases extracted are valid and human-interpretable.
\item Each phrase is assigned a topic in a principled manner.
\item The overall method is computationally efficient and of comparable complexity to LDA.
\item The topic model demonstrates similar perplexity to LDA
\end{itemize}

In addition to the above requirements for the system as a whole, we specify the requirements of a topic-representative phrase. When designing our phrase mining framework, we ensure that our phrase-mining and phrase-construction algorithms naturally validate candidate phrases on three qualities that constitute human-interpretability.

\begin{enumerate}
\item \textbf{Frequency:} The most important quality when judging whether a phrase relays important information regarding a topic is its frequency of use within the topic. A phrase that is not frequent within a topic, is likely not important to the topic. This formulation can be interpreted as a generalization of the list of most probable unigrams visualization used for LDA to a list of the most probable phrases one will encounter in a given topic.
\item \textbf{Collocation:} In corpus linguistics, a collocation refers to the co-occurence of tokens in such frequency that is significantly higher than what is expected due to chance. A commonly-used example of a phraseological-collocation is the example of the two candidate collocations ``strong tea" and ``powerful tea"\cite{halliday1966lexis}. One would assume that the two phrases appear in similar frequency, yet in the English language, the phrase ``strong tea" is considered more correct and appears in much higher frequency. Because a collocation's frequency deviates from what is expected, we consider them `interesting' and informative. This insight motivates the necessity of analyzing our phrases probabilistically to ensure they are collocations.
\item \textbf{Completeness:} If long frequent phrases satisfy the above criteria, then their subsets also satisfy these criteria. For example in the case of ``mining frequent patterns", ``mining frequent" will satisfy the frequency and collocation restriction, yet is clearly a subset of a larger and more intuitive phrase. Our phrase-construction algorithm should be able to automatically determine the most appropriate size for a human-interpretable phrase.
\end{enumerate}
We will introduce a framework that naturally embeds these phrase requirements.

\section{ToPMine Framework}
\label{sec:method}
To extract topical phrases that satisfy our desired requirements, we propose a framework that can be divided into two main parts: phrase-mining with text segmentation and phrase-constrained topic modeling. Our process for transforming a `bag-of-words'  document to a high-quality `bag-of-phrases' involves first mining frequent phrases, and then using these phrases to segment each document through an agglomerative phrase construction algorithm. After inducing a partition on each document, we perform topic modeling to associate the same topic to each word in a phrase and thus naturally to the phrase as a whole.

The goal of our phrase mining is to collect aggregate statistics for our phrase-construction. The statistical significance measure uses these aggregates to guide segmentation of each document into phrases. This methodology leverages phrase context and phrase significance in the construction process ensuring all phrases are of high-quality. The resultant partition is the input for our phrase-constrained topic model.

We choose the `bag-of-phrases' input over the traditional `bag-of-words' because under the latter assumption, tokens in the same phrase can be assigned to different latent topics. We address this by proposing a topic model PhraseLDA, which incorporates the `bag-of-phrases' partition from our phrase mining algorithm as constraints in the topical inference process. We have derived a collapsed Gibb's sampling method that when performing inference, ensures that tokens in the same phrase are assigned to the same topic. 
We expound upon our phrase-mining algorithm and topic model in Section~\ref{sec:phrase} and Section~\ref{sec:model} respectively.

\section{Phrase Mining}
\label{sec:phrase}
We present a phrase-mining algorithm that given a corpus of documents, merges the tokens within the document into human-interpretable phrases. Our method is purely data-driven allowing for great cross-domain performance and can operate on a variety of datasets. We extract high-quality phrases by obtaining counts of frequent contiguous patterns, then probabilistically reasoning about these patterns while applying context constraints to discover meaningful phrases.

The phrase mining algorithm can be broken down into two major steps. First, we mine the corpus for frequent candidate phrases and their aggregate counts. We have developed a technique that can quickly collect this information without traversing the prohibitively large search space. Second, we agglomeratively merge words in each document into quality phrases as guided by our \emph{significance measure}. We will discuss these steps in greater detail in the next two subsections.

\subsection{Frequent Phrase Mining}
In Algorithm~\ref{alg:FUCP Mining}, we present our frequent phrase mining algorithm. The task of frequent phrase mining can be defined as collecting aggregate counts for all contiguous words in a corpus that satisfy a certain minimum support threshold. We draw upon two properties for efficiently mining these frequent phrases.
\begin{enumerate}
\item Downward closure lemma: If phrase $P$ is not frequent, then any super-phrase of $P$ is guaranteed to be \textbf{not} frequent.
\item Data-antimonotonicity: If a document contains no frequent phrases of length $n$, the document does \textbf{not} contain frequent phrases of length $>$ $n$.
\end{enumerate}

The downward closure lemma was first introduced for mining general frequent patterns using the Apriori algorithm~\cite{agrawal1994fast}. We can exploit this property for our case of phrases by maintaining a set of active indices. These active indices are a list of positions in a document at which a contiguous pattern of length $n$ is frequent. In line 1 of Algorithm~\ref{alg:FUCP Mining}, we see the list of active indices. 

In addition, we use the data-antimonotonicity property to assess if a document should be considered for further mining~\cite{han2006data}. If the document we are considering has been deemed to contain no more phrases of a certain length, then the document is guaranteed to contain no phrases of a longer length. We can safely remove it from any further consideration. These two pruning techniques work well with the natural sparsity of phrases and provide early termination of our algorithm without searching through the prohibitively large candidate phrase space.

\begin{algorithm}[!th!]
\small
\Indm
\caption{Frequent Phrase Mining}\label{alg:FUCP Mining}
       \KwIn{Corpus with $D$ documents, min support $\epsilon$} 
       \KwOut{Frequent phrase and their frequency: \{(P,C(P))\}
       }
\Indp
       \BlankLine
		$\mathcal{D}\gets [D]$\\
	   	$A_{d,1} \gets \{\text{indices of all length-1 phrases} \in d\}\quad\forall d \in \mathcal{D}$   \\ 
	
       $C \gets HashCounter$(counts of frequent length-1 phrases)\\
	   $n \gets 2$\\
	   \While{ $\mathcal{D}\neq \emptyset$} {
		   \For{$d \in \mathcal{D}$}{
		   		\eat{$L_{n-1} \gets p=\{w_{d,i} .. w_{d,i+n-2} | i \in A_{d,n-1} \wedge C[p] \ge \epsilon \}$\\}
				$A_{d,n} \gets \{i\in A_{d,n-1} | C[\{w_{d,i} .. w_{d,i+n-2}\}]  \ge \epsilon \}$\\
				$A_{d,n} \gets A_{d,n}\setminus \{max(A_{d,n})\}$\\
				
				\eIf{$A_{d,n} = \emptyset$}{
				$\mathcal{D}\gets\mathcal{D}\setminus\{d\} $\\
					}{
						\For{$i \in A_{d,n}$}
						{
						\If{$i+1\in A_{d,n}$}
							{
							$P \gets \{w_{d,i} ..  w_{d,i+n-1}\}$\\
							 $C[P] \gets C[P] + 1$
							 }
						}
		  			}	
		   }
		   $n \gets n+1$
	  	 }
		 return $\{(P,C[P])| C[P]    \ge \epsilon\}$
\end{algorithm}
\normalsize

\eat{
\begin{algorithm}[h]
\Chi{Add some comments. Use Algorithm2e if there is time. Line ``LastIndex"}
\caption{Frequent Contiguous Pattern Mining}          
\label{FUCP Mining}    
\begin{algorithmic}
\Function{MinePatterns}{$Corpus, \epsilon$}
\State $A_{d1} \gets \{\text{indices of frequent 1-patterns} \in d\}$
\State $c \gets HashCounter$
\State $k \gets 2$
\While{NumDocs > 0} 
    \For{d $\in$ Corpus}
	\State $L_{k-1} \gets p=\{d_i .. d_{i+k-2} | i \in A_{d(k-1)} \wedge c[p] \ge \epsilon \}$
	\State $LastIndex \gets max(A_{d(k-1)})$
	\State $A_{dk} \gets \{i | c[\{d_i .. d_{i+k-2}|i \in A_{d(k-1)}]  \ge \epsilon \}$
	\State $A_{dk}.remove(LastIndex)$
	\If{$A_{dk} = \emptyset$}
		\State $NumDocs \gets NumDocs -1$
		\State Remove $d$ from consideration
	\Else
		\For{$i \in A_{dk}$}
		\State $P \gets \{d_i .. d_{i+k-1}\}$
		\State $c[P] \gets c[P] + 1$
		\EndFor
	\EndIf
    \EndFor
\EndWhile
    \State \Return $\bigcup_{k}L_k$, c
\EndFunction
\end{algorithmic}
\end{algorithm}
}
We take an increasing-size sliding window over the corpus to generate candidate phrases and obtain aggregate counts. At iteration $k$, for each document still in consideration, fixed-length candidate phrases beginning at each active index are counted using an appropriate hash-based counter. As seen in Algorithm~\ref{alg:FUCP Mining} line 7, candidate phrases of length $k-1$ are pruned if they do not satisfy the minimum support threshold and their starting position is removed from the active indices. We refer to this implementation of the downward closure lemma as \emph{position-based Apriori pruning}. As seen in Algorithm~\ref{alg:FUCP Mining} lines 9 - 11, when a document contains no more active indices, it is removed from any further consideration. This second condition in addition to pruning the search for frequent phrases provides a natural termination criterion for our algorithm.

The frequency criterion requires phrases to have sufficient occurrences. In general, we can set a minimum support that grows linearly with corpus size. The larger minimum support is, the more precision and the less recall is expected.

General frequent transaction pattern mining searches an exponential number of candidate patterns~\cite{agrawal1994fast,han2000mining}. When mining phrases, our contiguity requirement significantly reduces the number of candidate phrases generated. Worst case time-complexity occurs when the entire document under consideration meets the minimum support threshold. In this scenario,  for a document $d$ we generate $\mathcal{O}(\mathcal{N}_{d}^2)$ (a quadratic number) candidate phrases. Although this quadratic time and space complexity seems prohibitive, several properties can be used to ensure better performance. First, separating each document into smaller segments by splitting on phrase-invariant punctuation (commas, periods, semi-colons, etc) allows us to consider constant-size chunks of text at a time. This effectively makes the overall complexity of our phrase mining algorithm linear, $\mathcal{O}(N)$, in relation to corpus size. The downward closure and data antimonotonicity pruning mechanisms serve to further reduce runtime. 

\subsection{Segmentation and Phrase Filtering}
Traditional phrase extraction methods filter low quality phrases by applying a heuristic ``importance" ranking that reflect confidence in candidate key phrases, then only keeping the top-ranked phrases~\cite{mihalcea2004textrank}. Some methods employ external knowledge bases or NLP constraints to filter out phrases~\cite{liu2009clustering,mihalcea2004textrank}.

Our candidate phrase filtering step differentiates itself from traditional phrase extraction methods by implicitly filtering phrases in our document segmentation step. By returning to the context and constructing our phrases from the bottom-up, we can use phrase-context and the partition constraints to determine which phrase-instance was most likely intended. Because a document can contain at most a linear number of phrases (the number of terms in the document) and our frequent phrase mining algorithm may generate up to a quadratic number of candidate phrases, a quadratic number of bad candidate phrases can be eliminated by enforcing the partition constraint. 

The key element of this step is our bottom-up merging process. At each iteration, our algorithm makes locally optimal decisions in merging single and multi-word phrases as guided by a statistical significance score. In the next subsection, we present an agglomerative phrase-construction algorithm then explain how the significance of a potential merging is evaluated and how this significance guides our agglomerative merging algorithm.

\subsubsection{Phrase Construction Algorithm}
The main novelty in our phrase mining algorithm is the way we construct our high-quality phrases by inducing a partition upon each document. We employ a bottom-up agglomerative merging that greedily merges the best possible pair of candidate phrases at each iteration. This merging constructs phrases from single and multi-word phrases while maintaining the partition requirement. Because only phrases induced by the partition are valid phrases, we have implicitly filtered out phrases that may have passed the minimum support criterion by random chance. 

In Algorithm~\ref{alg:Document Partitioning}, we present the phrase construction algorithm. The algorithm takes as input a document and the aggregate counts obtained from the frequent phrase mining algorithm. It then iteratively merges phrase instances with the strongest association as guided by a potential merging's significance measure. The process is a bottom-up approach that upon termination induces a partition upon the original document creating a `bag-of-phrases'.

\begin{algorithm}[!th!]
\caption{Bottom-up Construction of Phrases from Ordered Tokens}
\label{alg:Document Partitioning}
\Indm
\small
       \KwIn{Counter $C$, thresh $\alpha$} 
       \KwOut{Partition}
\Indp
       \BlankLine
       $H \gets MaxHeap()$\\
       Place all contiguous token pairs into H with their significance score key.	\\

	   \While{H.size() $>$ 1} {
		   $Best \gets H.getMax()$\\
		   \eIf {Best.Sig $\ge$ $\alpha$}{
		   		$New \gets Merge(Best)$\\
				Remove Best from H\\
				Update significance for $New$ with its left phrase instance and right phrase instance\\
		   }{
			   $break$
		   }
	   }
\end{algorithm}
\normalsize
  
\eat{
\begin{algorithm}[h]
\caption{Bottom-up Construction of Phrases from Ordered Unigrams}          
\label{Document Partitioning}    
\begin{algorithmic}
\Function{Merge}{$d, L, \epsilon$}
    \State $H\gets MaxHeap()$
    \State Place all contiguous token pairs into H with their significance score key.
\While{H.size() > 1} 
	\State $Best \gets H.getMax()$
	\If {Best.Sig $\ge$ $\epsilon$}
		\State $New \gets Best \cup Best.RightPhrase$
		\State Remove Best.RightPhrase
		\State update significance for $New$ with its left phrase instance and right phrase instance
		\State Put $New$ in H
	\Else
		\State $break$
	\EndIf
\EndWhile
    \State \Return Partition
\EndFunction
\end{algorithmic}
\end{algorithm}
}

\begin{figure}[h]
\includegraphics[scale=0.41]{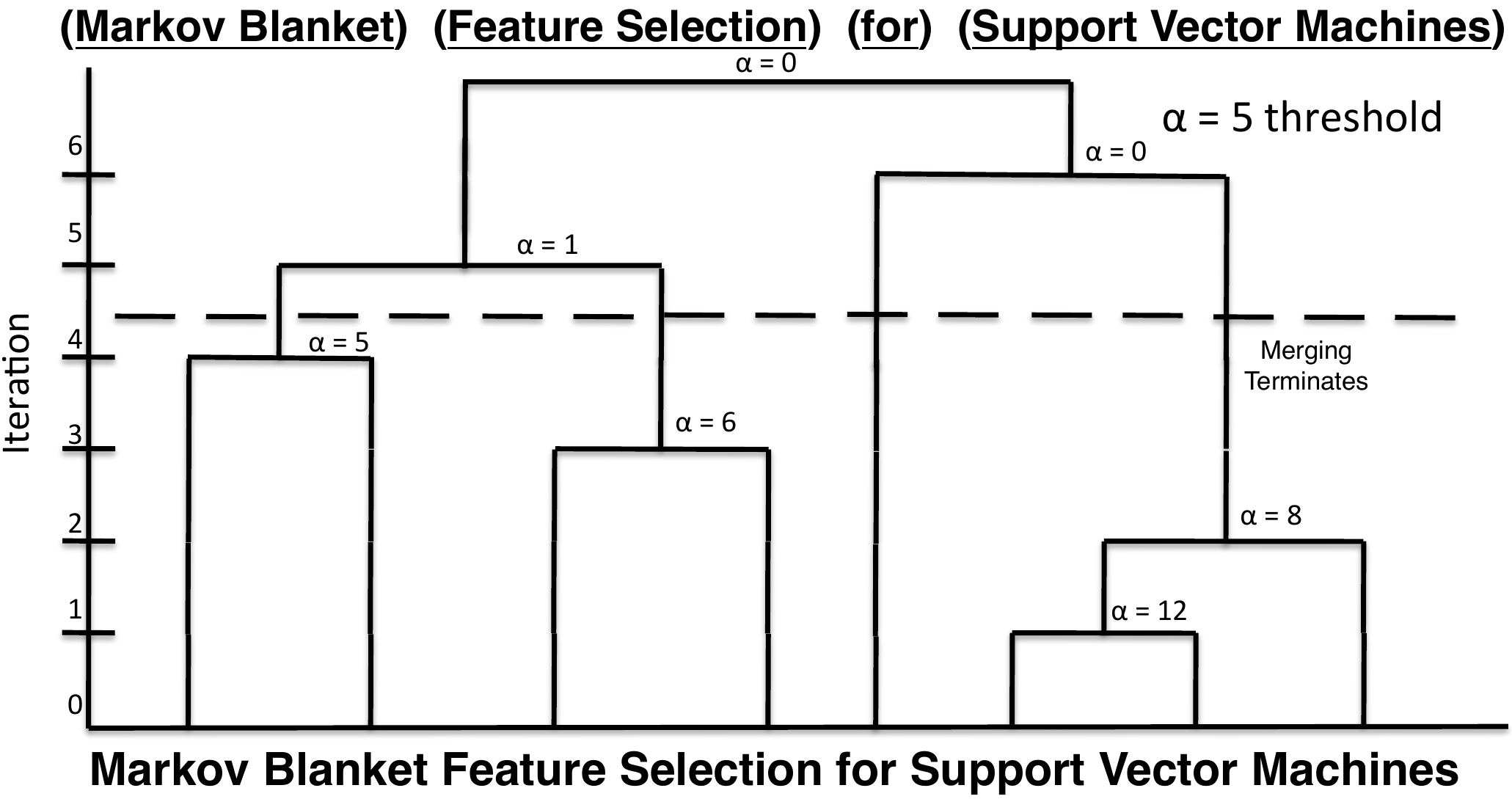}
\caption{Bottom-up construction of a `bag-of-phrases' on computer science title taken from DBLP.}
\label{fig:dendogram}
\end{figure}
Figure~\ref{fig:dendogram} tracks the phrase construction algorithm by visualizing the agglomerative merging of phrases at each iteration with a dendogram. Operating on a paper title obtained from our dblp titles dataset, each level of the dendogram represents a single merging. At each iteration, our algorithm selects two contiguous phrases such that their merging is of highest significance (Algorithm~\ref{alg:Document Partitioning} line 4) and merges them  (Algorithm~\ref{alg:Document Partitioning} lines 6 - 9) . The following iteration then considers the newly merged phrase as a single unit. By considering each newly merged phrase as a single unit and assessing the significance of merging two phrases at each iteration, we successfully address the ``free-rider" problem where long, unintelligible, phrases are evaluated as significant when comparing the occurrence of a phrase to the occurrence of each constituent term independently. 

As all merged phrases are frequent phrases, we have fast access to the aggregate counts necessary to calculate the significance values for each potential merging. By using proper data structures, the contiguous pair with the highest significance can be selected and merged in logarithmic time, $\mathcal{O}(log(\mathcal{N}_d))$ for each document. This complexity can once again be reduced by segmenting each document into smaller chunk by splitting on phrase-invariant  punctuation. Our algorithm terminates when the next merging with the highest significance does not meet a predetermined significance threshold $\alpha$ or when all the terms have been merged into a single phrase. This is represented by the dashed line in Figure~\ref{fig:dendogram} where there are no more candidate phrases that meet the significance threshold. Upon termination, a natural ``bag-of-phrases" partition remains. While the frequent phrase mining algorithm satisfies the \emph{frequency} requirement, the phrase construction algorithm satisfies the collocation and completeness criterion.
\eat{
\subsubsection{Significance Measure}
  \begin{table}[h]
  \centering
  \caption{Significance Measure Notation}
    \begin{tabular}{|c|c|}
    \hline
    \textbf{Notation} & \textbf{Interpretation} \bigstrut\\
    \hline
    N     & Total number terms in corpus \bigstrut\\
    \hline
    $p(P_x)$     & Actual probability of phrase x \bigstrut\\
    \hline
    $\left\vert{phr_x}\right\vert$     & Actual frequency of phrase x \bigstrut\\
    \hline
    $h_{0}$     & Null hypothesis (assumes independence)\bigstrut\\
    \hline
    $\mu_{0}$     & Probability of phrase x under $h_0$ \bigstrut\\
    \hline
    $s_x^2$		& Sample variance of phrase x  \bigstrut\\
    \hline 
    $phr_{x+y}$		& phrase x $\cup$ phrase y \bigstrut\\
    \hline
    \end{tabular}%
  \label{tab:addlabel}%
\end{table}%
}

To statistically reason about the occurrence of phrases, we consider a null hypothesis, that the corpus is generated from a series of independent Bernoulli trials. Under this hypothesis, the presence or absence of a phrase at a specific position in the corpus is a product of a Bernoulli random variable, and the expected number of occurrences of a phrase can be interpreted as a binomial random variable. Because the number of tokens $L$ in the corpus can be assumed to be fairly large, this binomial can be reasonably approximated by a normal distribution. As such, the null hypothesis distribution, $h_0$, for the random variable $f(P)$, the count of a phrase $P$ within the corpus is:

 \begin{align*}
 h_0(f(P)) &= \mathcal{N}(Lp(P), Lp(P)(1-p(P))) \\
 &\approx  \mathcal{N}(Lp(P), Lp(P)) 
 \end{align*}
 
where $p(P)$ is the Bernoulli trial success probability for phrase $P$. The empirical probability of a phrase in the corpus can be estimated as $p(P) = \frac{f(P)}{L}$.
 Consider a longer phrase that composed of two phrases $P_1$ and $P_2$. The mean of its frequency under our null hypothesis of independence of the two phrases is:
\begin{center}
$\mu_{0}(f(P_1\oplus P_2)) =  Lp(P_1) p(P_2)$
\end{center}
This expectation follows from treating each phrase as a constituent, functioning as a single unit in the syntax. Due to the unknown population variance and sample-size guarantees from the minimum support, we can estimate the variance of the population using sample variance: $\sigma_{P_1\oplus P_2}^2 \approx f(P_1\oplus P_2)$, the sample phrase occurrence count.
We use a significance score to provide a quantitative measure of which two consecutive phrases form the best collocation at each merging iteration. This is measured by comparing the actual frequency with the expected occurrence under $h_0$.
\begin{equation}
\label{eq:significance}
sig(P_1,P_2) \approx  \frac{f(P_1\oplus P_2)-  \mu_0(P_1,P_2)}{\sqrt{f(P_1\oplus P_2)}}
\end{equation}
Equation~\ref{eq:significance} computes the number of standard deviations away from the expected number of occurrences under the null model. This significance score can be calculated using the aggregate counts of the candidate phrases, which can be efficiently obtained from the frequent phrase-mining algorithm. This significance score can be considered a generalization of the t-statistic which has been used to identify dependent bigrams  \cite{church19916,pedersen1996fishing}. By checking the $h_0$ of merging two contiguous sub-phrases as opposed to merging each individual term in the phrase, we effectively address the `free-rider' problem where excessively long phrases appear significant. To address the concern that the significance score relies on the naive independence assumption, we do not perform hypothesis testing to accept or reject $h_0$. Instead we use the score as a robust collocation measure by which to guide our algorithm in selecting phrases to merge. A high significance indicates a high-belief that two phrases are highly associated and should be merged.

\eat{
\subsection{Phrase Mining Complexity Analysis}
The overall complexity of segmenting our input corpus into phrases depends on the frequent contiguous pattern mining for collection of aggregate statistics and the bottom-up agglomerative merging for segmentation. 
Frequent contiguous pattern mining differs from transaction pattern mining\cite{agrawal1994fast,han2000mining}  in that the worst-case complexity is quadratic and not exponential. This is because we are searching the state space for a constrained set of frequent patterns where the elements must be contiguous. To further reduce the complexity of mining these patterns, we exploit the natural boundary structure often found in text data - punctuation. Splitting on phrase-invariant punctuation, such as periods, commas, and semicolons, allows us to consider small chunks at a time when searching for frequent phrases. Because these chunks rarely exceed a constant number of words, splitting on this punctuation yields a linear-time algorithm in relation to corpus size. As such our frequent contiguous pattern mining algorithm is $\mathcal{O}(N)$ in relation to corpus size. Our Apriori pruning and data-antimonotonicity heuristics provide a reduction in the leading coefficient reducing overall runtime and allowing for early termination. 

While each selection of candidate phrases for merging is done in logarithmic time, once again splitting on phrase-invariant punctuation reduces the logarithmic time selection to constant time. As such the total runtime of the phrase construction algorithm when splitting on punctuation is performed is once again $\mathcal{O}(N)$. With both phases of the text segmentation processes showing linear time complexity, our complete phrase mining algorithm possesses linear time complexity. This is experimentally verified in the Experimental Results section.
}
\section{Topic modeling}
\label{sec:model}
In the previous section, we segment a document into a collection of phrases, which provides a new representation for documents, i.e. `bag-of-phrases'. These phrases are a group of words that appear frequently, contiguously, and occur more often than due to chance. Our insight is that \textit{with high probability}, tokens in the same phrase should share the same latent topic.

In this section, we start with a brief review of Latent Dirichlet Allocation~\cite{blei2003latent}, and then propose a novel probabilistic model, PhraseLDA, which incorporates the `constraint' idea into LDA. A collapsed Gibbs sampling algorithm is developed for PhraseLDA, and optimization of hyper-parameters is discussed. Finally, we define topical frequency for phrases, which serves as a ranking measure for our topic visualization. We list the notations used in Table~\ref{tab:topmod}, where $\mathcal{I}(statement) =1$ if statement is true; otherwise 0. We denote $Z$ the collection of all latent variables $\{z_{d,g,j}\}$, and $W,\Theta,\Phi$ the collection of their corresponding random variables $\{w_{d,g,j}\},\{\theta_d\},\{\phi_k\}$.

\begin{table}[h!]
\centering
  \caption{Notation used in topic modeling}
      \small
    \begin{tabular}{|p{1.5cm}|p{6.5cm}|}
    \hline
    \textbf{Variable} & \textbf{Description} \bigstrut\\
    \hline
    D, K, V     &  number of documents, topics, size of vocabulary\bigstrut \\
    \hline
    d, g, i, k, x     & index for document, phrase in a doc, token in a doc, topic, word \bigstrut \\
    \hline
    $N_d$ & number of tokens in  d\textit{th} doc \bigstrut \\
    \hline
     $G_d$ & number of phrases in d\textit{th} doc(after partition)  \bigstrut\\
    \hline
    $W_{d,g}$ & number of tokens in  g\textit{th} phrase  of d\textit{th} doc \bigstrut \\
    \hline
    $\theta_{d}$ & multinomial distribution over topics for d\textit{th} doc \bigstrut \\
    \hline
    $z_{d,g,j}$ &  latent topic for j\textit{th} token in g\textit{th} phrase  of d\textit{th} doc \bigstrut \\
    \hline
    $w_{d,g,j}$ & the j\textit{th} token in g\textit{th} phrase  of doc d \bigstrut \\
    \hline
    $\phi_{k}$ & multinomial distribution over words in topic k \bigstrut \\
    \hline
    $\mathcal{N}_k$ & $\mathcal{N}_k = \sum_{d,g,j} \mathcal{I}(z_{d,g,j} == k)$, number of tokens assigned to topic k \bigstrut \\
    \hline
    $\mathcal{N}_{d,k}$ & $\mathcal{N}_{d,k} = \sum_{g,j} \mathcal{I}(z_{d,g,j} ==k)$, number of tokens assigned to topic k in doc d \bigstrut \\
     \hline
    $\mathcal{N}_{x,k}$ & $\mathcal{N}_{x,k} = \sum_{d,g,j} \mathcal{I}(z_{d,g,j} ==k, w_{d,g,j}==x)$, number of tokens with value x and topic k  \bigstrut \\
    \hline
    $\alpha, \beta$ & parameter of the Dirichlet distribution for $\theta_d,\phi_k$ \bigstrut \\
    \hline
    $\mathcal{C}_{d,g}$ & $\{z_{d,g,j}\}_{j=1}^{W_{d,g}}$, the collection of all latent variables in g\textit{th} clique(phrase) of d\textit{th} doc \bigstrut \\
    \hline
    \end{tabular}%
  \label{tab:topmod}%
\end{table}%

\subsection{Brief review of LDA}
LDA assumes that a document is a mixture of topics, where a topic is defined to be a multinomial distribution over words in the vocabulary. The generative process is as follows:
\begin{enumerate}
\item Draw $\phi_k \sim Dir(\beta)$, for $k = 1,2,...,K$
\item For d\textit{th} document, where $d = 1,2,...,D$:
\begin{enumerate}
\item Draw $\theta_d \sim Dir(\alpha)$
\item For i\textit{th} token in d\textit{th} document, where $i = 1,2,...,N_d$:
\begin{enumerate}
\item Draw $z_{d,i} \sim Multi(\theta_d)$ 
\item Draw $w_{d,i} \sim Multi(\phi_{z_{d,i}})$ 
\end{enumerate}
\end{enumerate}
\end{enumerate}
\begin{figure}
\centering
\subfigure[\small Bayesian network for LDA]{
\includegraphics[width=0.15\textwidth]{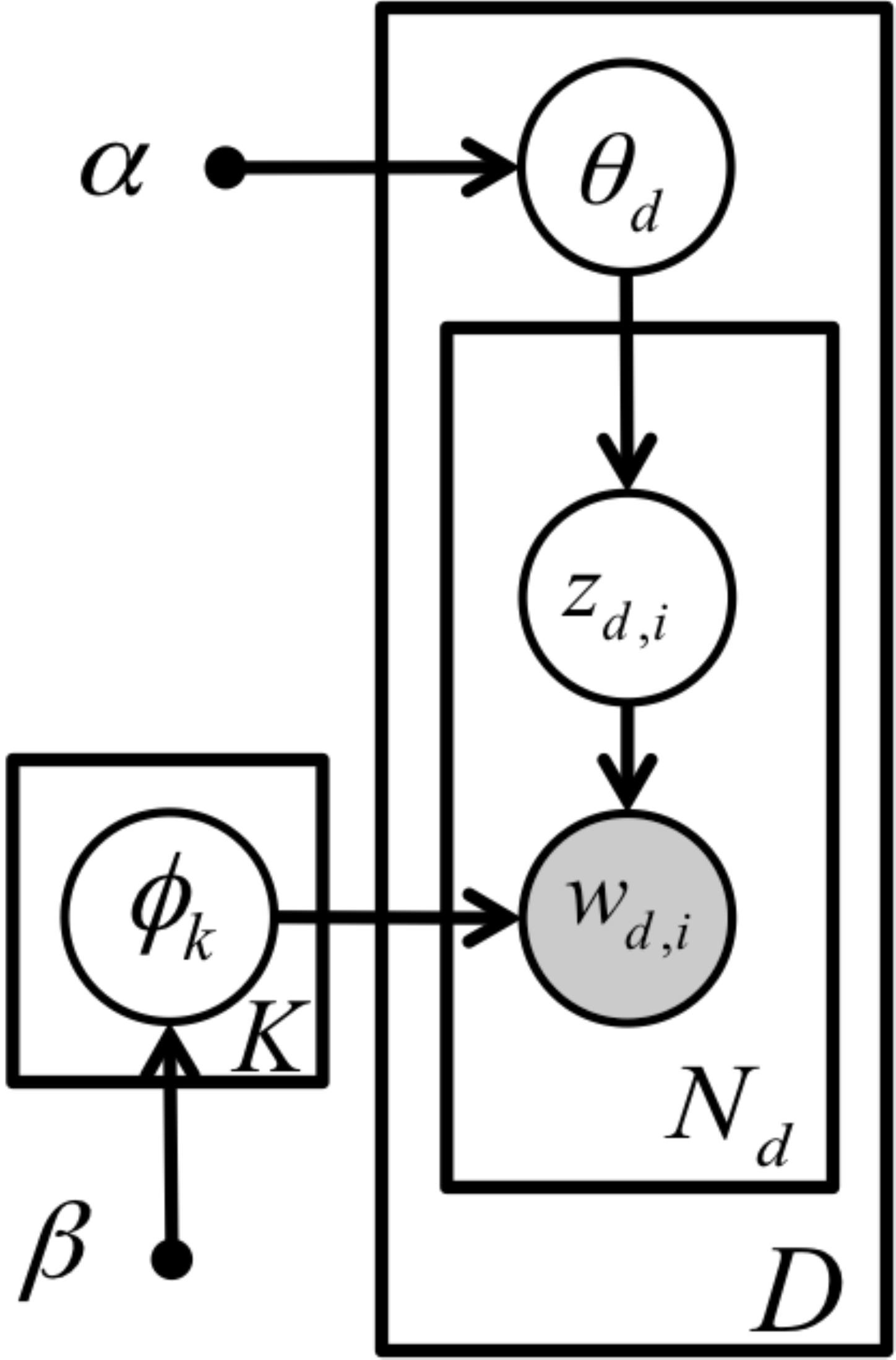}
\label{fig:LDA}}
\hspace{0.05in}
\subfigure[\small chain graph for PhraseLDA]{
\centering
\includegraphics[width=0.27\textwidth]{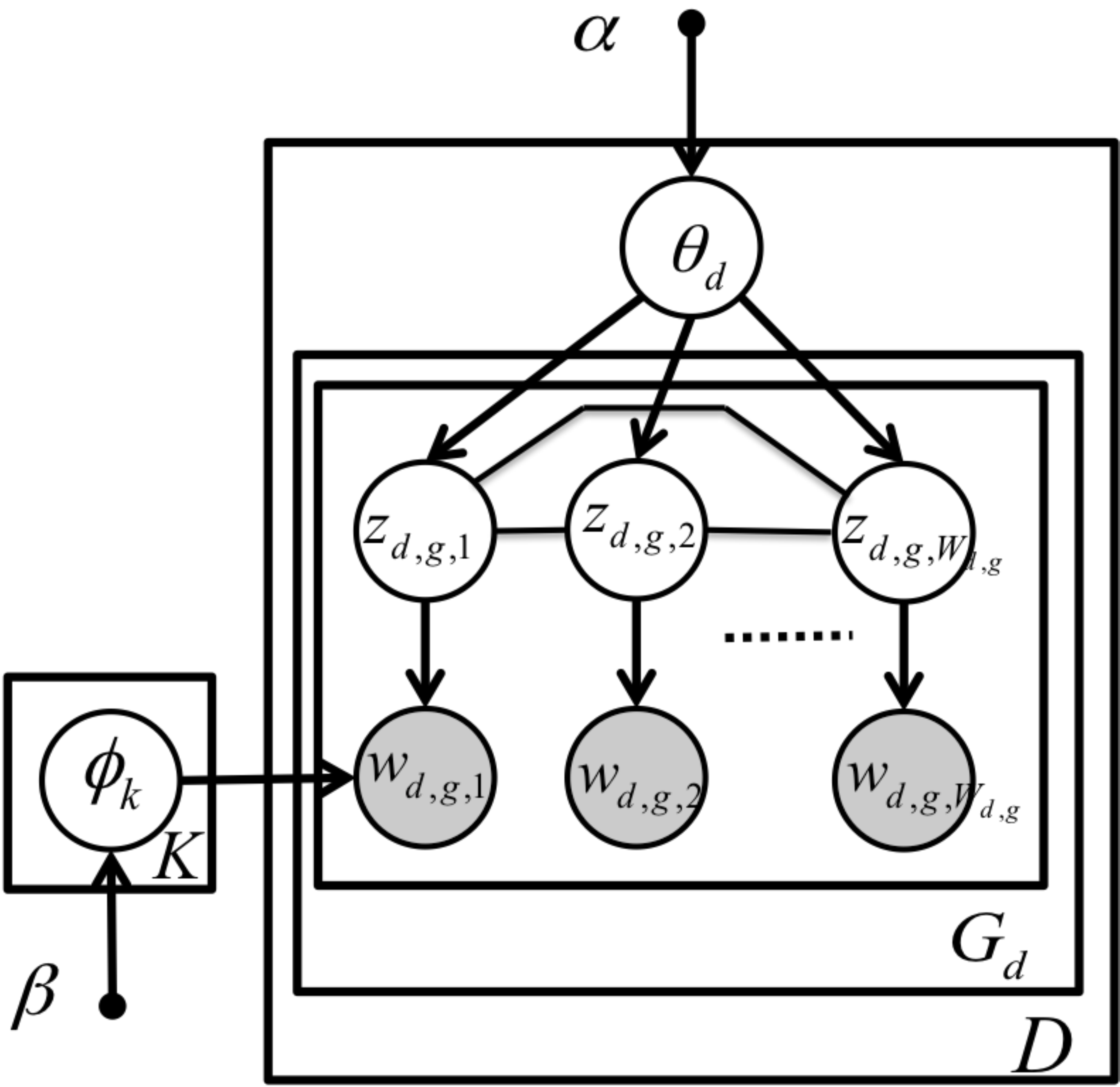}
\label{fig:histogram}}
\caption{\scriptsize In PhraseLDA, latent topic variables ($z_{d,g,j}$) in the same phrase form a clique. Each clique introduces a potential function onto the joint distribution defined by LDA. For computational efficiency, we choose a potential function which assigns same-clique tokens to the same topic}
\label{fig:phraseLDA}
\end{figure}

The graphical model for LDA, depicted in Figure~\ref{fig:LDA}, defines the joint distribution of random variables. By utilizing the conditional independence encoded in the graph, the joint distribution can be written as(we omit the hyper-parameter $\alpha,\beta$ for simplicity):
\begin{equation}
P_{\text{\small{LDA}}}(Z,W,\Phi,\Theta) = \prod_{d,i} p(z_{d,i}|\theta_d)p(w_{d,i}|z_{d,i},\Phi) \prod_{d} p(\theta_d) \prod_{k}p(\Phi_k)
\end{equation}

Because of the conjugacy between multinomial and Dirichlet distributions, we can easily integrate out $\{\Theta,\Phi\}$. That is,
\begin{equation}
P_{\text{\small{LDA}}}(Z,W) = \int P_{\text{\small{LDA}}}(Z,W,\Phi,\Theta)  d\Theta d\Phi
\end{equation}
has a closed form (see Appendix).
\newline
\newline
\subsection{PhraseLDA}
LDA is built upon the `bag-of-words' assumption, under which the order of words is completely ignored. As a result, when inferring the topic assignment $z_{d,i}$ for word $w_{d,i}$, the topic of a far-away word in the same document has the same impact as a near-by word. In section~\ref{sec:phrase}, we partition a document into a collection of phrases.\eat{A phrase contains contiguous words, which appear in the corpus much more often than due to chance.} We believe that the high frequency of occurrence significantly greater than due to chance, and the proximity constraint induced by contiguity are an indication that there is a stronger correlation than that expressed by LDA between the words in a phrase.  This motivates our use of the chain graph to model this stronger correlation. Chain graphs are most appropriate when there are both response-explanatory relations (Bayesian networks) and symmetric association relations (Markov networks) among variables~\cite{ma2008structural}. In our task, LDA \eat{, which is a generative model,} models the (directed) causal relations between topics and the observed tokens,\eat{ but due to its `bag-of-words' assumption, ignores the fact that near-by words have stronger correlation than far-away words.} and we propose to use un-directed graph to model the stronger dependence among near-by words.

We connect the latent topic assignments in the same phrase using un-directed edges(as shown in figure~\ref{fig:phraseLDA}). As a result, for g\textit{th} phrase of d\textit{th} document, random variables $\{z_{d,g,j}\}_{j=1}^{W_{d,g}}$ form a clique. 

Every clique $\mathcal{C}_{d,g}$(or equivalently phrase) introduces a potential function $f(\mathcal{C}_{d,g})$, which should express the intuition that with high probability, $z_{d,g,j}$'s in the clique should take the same value. As a result, the chain graph defines the joint distribution over all random variables:
\begin{equation}
P(Z,W,\Phi,\Theta) = \frac{1}{C} P_{\text{\small{LDA}}}(Z,W,\Phi,\Theta) \prod_{d,g} f(\mathcal{C}_{d,g})
\label{equ:joint}
\end{equation}
where $C$ is a normalizing constant that makes the left hand side a legitimate probability distribution.

\subsection{Inference} \label{sec:phraseLDA}
 For the joint distribution defined by equation~\ref{equ:joint}, we developed a collapsed gibbs sampling algorithm to sample latent assignment variables $Z$ from its posterior. As with LDA, the first step is to integrate out $\{\Theta,\Phi\}$:
\begin{align}
P(Z,W)& = \int  \frac{1}{C} P_{\text{\small{LDA}}}(Z,W,\Phi,\Theta) \prod_{d,g} f(\mathcal{C}_{d,g}) \; d\Theta d\Phi \nonumber \\ 
& = \frac{1}{C} \;\left(\int  P_{\text{\small{LDA}}}(Z,W,\Phi,\Theta) d\Theta d\Phi \right)  \prod_{d,g} f(\mathcal{C}_{d,g}) \; \nonumber \\
&= \frac{1}{C} \;P_{\text{\small{LDA}}}(Z,W) \prod_{d,g} f(\mathcal{C}_{d,g}) \label{equ:collopsed}
\end{align}
$P(Z,W)$ takes a simple closed form because $P_{\text{\small{LDA}}}(Z,W)$ does.

Ideally, the potential function in equation~\ref{equ:joint} expresses the strong (symmetric) influence between the words within a phrase. \eat{However a general potential function is computationally intractable.} Suppose clique $\mathcal{C}_{d,g}$ is of size s, then  $\mathcal{C}_{d,g}$ can be in any of the $K^s$ possible states, where $K$ is the number of topics. Since the normalizing constant is unknown, we need to compute a value for all  $K^s$ states, and then normalize the values to get a legitimate distribution, which is computationally intractable for large $K$ and $s$. As such, we choose a specific potential function below:
\begin{equation}
f(\mathcal{C}_{d,g}) = 
  \begin{cases}
   \;1 & \text{if} \quad z_{d,g,1} = z_{d,g,2}=...=z_{d,g,W_{d,g}} \\
   \;0     & \text{otherwise}
  \end{cases}
  \label{equ:potential}
\end{equation}

This potential function coerces all variables in the clique to take on the same latent topic. Because our phrase-mining algorithm performs a constrained merging guided by a statistical significance measure, we assume that it is of high probability that the random variables in the clique possess the same topic. As such, we adopt the potential function as specified by equation~\ref{equ:potential} as an approximation, which reduces the possible states of $\mathcal{C}_{d,g}$ from $K^s$ to $K$. Next, we  develop an efficient gibbs sampling algorithm for this particular choice. 

\eat{Our justification is that our phrase-mining algorithm performs a constrained merging guided by a statistical significance measure to ensure all phrases mined are collocations. By setting a high significance threshold for termination, we can ensure parsimony in what we constitute a phrase almost guaranteeing that each of our mined phrases should be construed as a single unit. We believe a single unit's constituents should certainly possess the same topic.
For the joint distribution defined by equation~\ref{equ:joint}, we developed a collapsed gibbs sampling algorithm to sample latent assignment variables $Z$ from its posterior. As with LDA, the first step is to integrate out $\{\Theta,\Phi\}$:
\begin{align*}
P(Z,W)& = \int  \frac{1}{C} P_{\text{\small{LDA}}}(Z,W,\Phi,\Theta) \prod_{d,g} f(\mathcal{C}_{d,g}) \; d\Theta d\Phi \nonumber \\ 
& = \frac{1}{C} \;\left(\int  P_{\text{\small{LDA}}}(Z,W,\Phi,\Theta) d\Theta d\Phi \right)  \prod_{d,g} f(\mathcal{C}_{d,g}) \; \nonumber \\
&= \frac{1}{C} \;P_{\text{\small{LDA}}}(Z,W) \prod_{d,g} f(\mathcal{C}_{d,g}) \label{equ:collopsed}
\end{align*}
$P(Z,W)$ takes a simple analytical form because $P_{\text{\small{LDA}}}(Z,W)$ does.
}
We sample a configuration for a clique $\mathcal{C}_{d,g}$ from its posterior $p(\mathcal{C}_{d,g}| W , Z_{\setminus \mathcal{C}_{d,g}})$. \eat{As we discussed above, we adopt the potential function given in equation~\ref{equ:potential}. As such,}Since $\mathcal{C}_{d,g}$ can only take $K$ possible configuration, we use $\mathcal{C}_{d,g} = k$ to indicate that  all variables in clique $\mathcal{C}_{d,g}$ taking value $k$. We show in the Appendix that
\small
\begin{align}
&p(\mathcal{C}_{d,g} = k |W , Z_{\setminus \mathcal{C}_{d,g}} ) \;\propto \nonumber\\
&\prod_{j=1}^{W_{d,g}} 
 \left(\alpha_{k} + \mathcal{N}_{d,k\setminus C_{d,g}}+j-1\right)
 \frac{
\left( \beta_{w_{d,g,j}} + \mathcal{N}_{w_{d,g,j},k\setminus \mathcal{C}_{d,g}} \right)
}{
\left(\sum_{x=1}^{V} \beta_x + \mathcal{N}_{k\setminus \mathcal{C}_{d,g}} + j - 1\right)
}
\end{align}
\normalsize

For a ``legitimate" $Z$, where the variables in the same clique take the same value, 
$p(Z,W|\alpha,\beta) = \frac{1}{C} P_{\text{\small{LDA}}}(Z,W|\alpha,\beta)$, which shows we can adopt the same hyper-parameter($\alpha,\beta$) optimization techniques as in LDA. In the experiment, we use the fixed-point method proposed by~\cite{minka2000estimating}.

\subsection{Topic visualization}

There is a large literature in ranking terms and phrases for effective topical visualization. One method for selecting representative phrases (label) for a topic can be to minimize Kullback-Leibler divergence between word distributions and maximizing mutual information between label phrases and the topic model\cite{mei2007automatic}. Another method attempts to extend the list of potential labels using external sources, such as wikipedia, and rank based on augmented candidate pool \cite{lau2011automatic}. Other methods provide a parameterized multi-faceted ranking function that allows for a more user-controlled ranking\cite{danilevsky2014kert}. These methods all provide a suggested methodology for ranking phrases within a topic model and can be easily incorporated into ToPMine.

For a more simplistic ranking function, we generalize the concept of N-most-probable terms in unigram LDA to our phrase output. By adopting the potential function given in equation~\ref{equ:potential}, all variables in a clique are guaranteed to have the same latent topic. Since a clique corresponds to a phrase, naturally we assign the phrase to the same topic as shared by its constituents. 

\eat{Denote $\text{PI}_{d,g}$  the g\textit{th} phrase instance in d\textit{th} documents.} We utilize the topic assignment for each token from the last iteration of gibbs sampling, and define topical frequency (TF) for a phrase \textit{phr} in topic k as the number of times it is assigned to topic k:
\begin{equation}
\text{TF(\textit{phr},k)} = \sum_{d,g} \mathcal{I}(\text{PI}_{d,g} == phr, C_{d,g} == k)
\end{equation}
where $\mathcal{I}(\cdot)$ is the indicator function as used before, and $\text{PI}_{d,g}$  is the g\textit{th} phrase instance in d\textit{th} documents.

With this definition, we can visualize topic $k$ by sorting the phrases according to their topical frequency in topic $k$. 

\section{Related Work}
\label{sec:related}

Recently many attempts have been made to relax the `bag-of-words' assumption of LDA. These topical phrase extraction techniques fall into two main categories, those that infer  phrases and topics simultaneously by creating complex generative models and those that apply topical phrase discovery as a post-process to LDA.

Methods have experimented with incorporating a bigram language model into LDA~\cite{wallach2006topic}. This method uses a hierarchical dirichlet to share the topic across each word within a bigram. TNG~\cite{wang2007topical} is a state-of-the-art approach to n-gram topic modeling that uses additional latent variables and word-specific multinomials to model bi-grams. These bigrams can be combined to form n-gram phrases. PD-LDA uses a hierarchal Pitman-Yor process to share the same topic among all words in a given n-gram ~\cite{lindsey2012phrase}. Because PD-LDA  uses a nonparametric prior to share a topic across each word in an n-gram, it can be considered a natural generalization of the LDA bigram language model to n-grams and more appropriate for comparison.

Other methods construct topical phrases as a post-processing step to LDA and other topic models. KERT constructs topical phrases by performing unconstrained frequent pattern mining on each topic within a document then ranking the resultant phrases based on four heuristic metrics~\cite{danilevsky2014kert}. Turbo Topics uses a back-off n-gram model and permutation tests to assess the significance of a phrase~\cite{blei2009visualizing}. 

Topical key phrase extraction has even been applied to the social networking service Twitter~\cite{zhao2011topical}. Using a Twitter-specific topic model and ranking candidate key phrases with an extension of topical page-rank on retweet behavior, the method extracts high-quality topical keywords from twitter. Because this method relies on the network topology of twitter, it doesn't extend to other text corpora. 

Attempts to directly enrich the text corpora with frequent pattern mining to enhance for topic modeling have also been investigated~\cite{kim2012enriching}. As the objective of this method is to enrich the overall quality of the topic model and not for the creation of interpretable topical phrases, their main focus is different from ToPMine.

The concept of placing constraints into LDA has been investigated in several methods. Hidden Topic Markov Model makes the assumption that all words in a sentence have the same topic with consecutive sentences sharing the same topic with a high probability~\cite{gruber2007hidden}. By relaxing the independence assumption on topics, this model displays a drop in perplexity while retaining computational efficiency. Sentence-LDA, is a generative model with an extra level generative hierarchy that assigns the same topic to all the words in a single sentence~\cite{jo2011aspect}. In both of these models, the final output produced is a general topic model with no intuitive method of extracting topical phrases. 

There is a large literature on unsupervised phrase extraction methods. These approaches generally fall into one of a few techniques: language modeling, graph-based ranking, and clustering~\cite{hasan2010conundrums}. Because these methods simply output a ranked list of phrases, they are incompatible with our phrase-based topic modeling which operates on partitioned documents. 

\section{Experimental results}
\label{sec:exp}
In this section, we start with the introduction of  the datasets we used and methods for comparison. We then describe the evaluation on interpretability and scalability.
\ \\

\newpage
\subsection{Datasets and methods for comparison} 
\noindent
\textbf{Datasets}
\ \\
We use the following six datasets for evaluation purpose:
\begin{itemize}
\item \textbf{DBLP}\eat{\footnote{\small\url{http://www.dblp.org}}} \textbf{titles}. We collect a set of titles of recently published computer science papers. The collection has 1.9M titles, 152K unique words, and 11M tokens.
\item  \textbf{20Conf}.  Titles of papers published in 20 conferences related to the areas of Artificial Intelligence, Databases, Data Mining, Information Retrieval, Machine Learning, and Natural Language Processing - contains 44K titles, 5.5K unique words, and 351K tokens.
\item \textbf{DBLP abstracts}. Computer science abstracts containing 529K abstracts, 186K unique words, and 39M tokens.
\item \textbf{TREC AP news}. News dataset(1989) containing 106K full articles, 170K unique words, and 19M tokens. 
\item \textbf{ACL abstracts}. ACL\eat{\footnote{\small\url{www.aclweb.org}}} abstracts containing 2k abstracts, 4K unique words and 231K tokens.
\item \textbf{Yelp Reviews}. Yelp reviews\eat{\footnote{\small\url{https://www.yelp.com/academic_dataset}}} containing 230k Yelp reviews and 11.8M tokens.
\end{itemize}

We perform stemming on the tokens in the corpus using the porter stemming algorithm\cite{porter1980algorithm} to address the various forms of words (e.g. cooking, cook, cooked) and phrase sparsity. We remove English stop words for the mining and topic modeling steps. Unstemming and reinsertion of stop words are performed post phrase-mining and topical discovery.

There are four directly comparable methods outlined in Section~\ref{sec:related}: Turbo Topics, TNG, PD-LDA, and KERT.
\eat{
\noindent
\textbf{Methods for Comparison.} 
\ \\
\indent We compare ToPMine with the following four methods:

\noindent
$\bullet$ TNG~\cite{wang2007topical} - A state-of-the art approach to n-gram topic modeling. By using additional latent variables to model bi-grams and adding word-specific multinomials, TNG can be used to construct topical phrases. 
\ \\
$\bullet$ TurboTopics~\cite{blei2009visualizing} - A post-processing algorithm to LDA. It leverages permutation tests and a back-off N-Gram language model to recursively merge same-topic terms from LDA into more understandable groupings.
\ \\
$\bullet$ KERT~\cite{danilevsky2013kert} - A topical key phrase extraction method that draws upon the the speed of general frequent pattern mining algorithms to mine frequent unordered patterns in the same document. These patterns are filtered and ranked under four criteria to create a list of topical phrases of mixed lengths.
\ \\
$\bullet$ PD-LDA~\cite{lindsey2012phrase} - A hierarchical topical model that infers both phrases and topics. Using hierarchical Pitman-Yor processes, the topic is naturally shared to all constituents in a phrase.
}
\subsection{Interpretability}
We propose two user studies to demonstrate the effectiveness of our ToPMine framework.

\subsubsection*{Phrase Intrusion}
First, we use an \emph{intrusion detection} task which adopts the idea proposed by~\cite{Chang:Boyd-Graber:Wang:Gerrish:Blei-2009} to evaluate topical separation. The intrusion detection task involves a set of questions asking humans to discover the `intruder' object from several options.\eat{Three annotators manually completed each task, and their evaluations scores were pooled. The results of this task evaluate how well the phrases are separated in different topics.} Each question consists of $4$ phrases; $3$ of them are randomly chosen from the top $10$ phrases of one topic and the remaining phrase is randomly chosen from the top phrases of a different topic. Annotators are asked to select the intruder phrase, or to indicate that they are unable to make a choice. The results of this task evaluate how well the phrases are separated in different topics

\begin{figure}[h]
\centering
\includegraphics[width=0.41\textwidth]{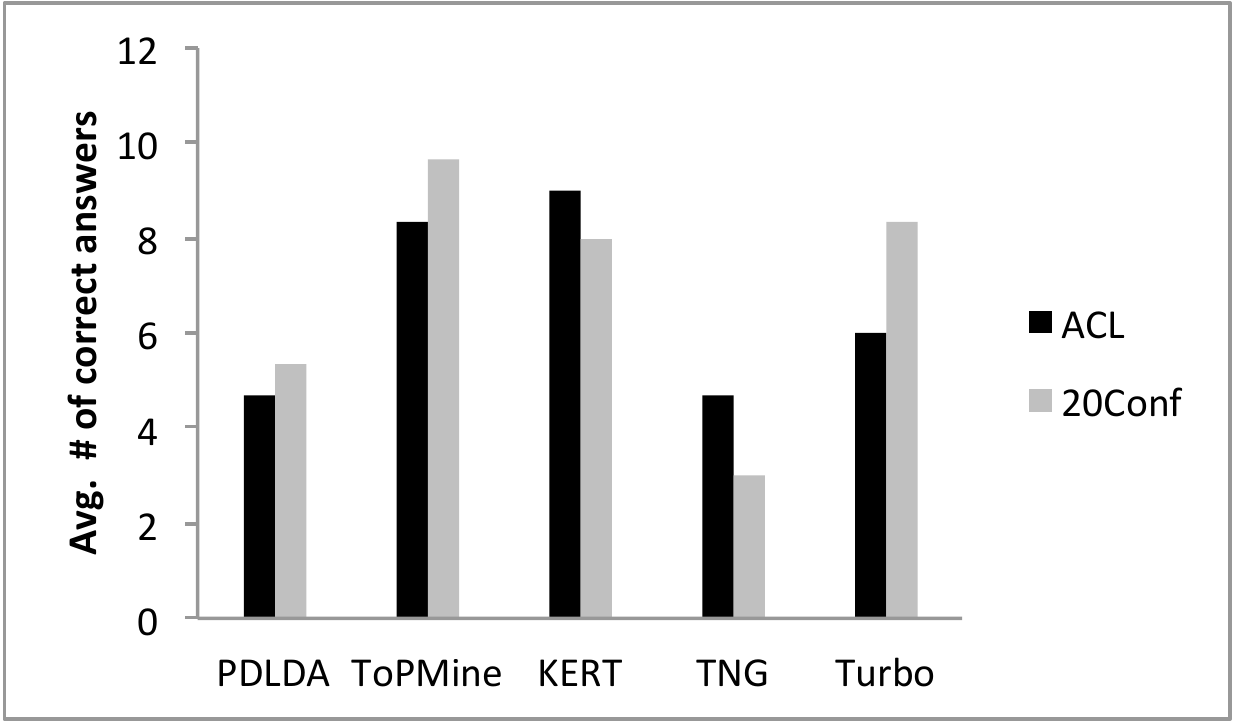}
\caption{Phrase intrusion task. Test subjects were asked to identify an intruder phrase in a topic.}
\label{fig:intrusion}
\end{figure}

For each method, we sampled 20 Phrase Intrusion questions, and asked three annotators to  answer each question. We report the average number of questions that is answered `correctly' (matching the method) in Figure~\ref{fig:intrusion}.

\eat{The results show KERT and ToPMine demonstrating similar performance for phrase intrusion on the  ACL and 20Conf  datasets. Turbo Topics demonstrates superior performance 
We then calculate the agreement of the human choices with the actual results returned by the various methods. We consider a higher match between a given method and human judgment to imply a higher quality. For each method, we report the average percent of questions answered `correctly' (matching the method), as well as the average percent of questions that annotators were able to answer. The results can be seen in Figure~\ref{fig:intrusion}.}

\subsubsection*{Domain Expert Evaluation}
The second task is motivated by our desire to extract high-quality topical phrases and provide an interpretable visualization. This task evaluates both topical coherence on the full topical phrase list and phrase quality. We first visualize each algorithm's topics with lists of topical phrases sorted by topical frequency. For each dataset, five domain experts (computer science and linguistics graduate students) were asked to analyze each method's visualized topics and score each topical phrase list based on two qualitative properties:

\begin{itemize}
\item \textbf{Topical coherence:} We define topical coherence as homogeneity of a topical phrase list's thematic structure. This homogeneity is necessary for interpretability. We ask domain experts to rate the coherence of each topical phrase list on a scale of $1~to~10$.
\item \textbf{Phrase quality:} To ensure that the phrases extracted are meaningful  and not just an agglomeration of words assigned to the same topic, domain experts are asked to rate the quality of phrases in each topic from $1~to~10$.
\end{itemize}

\begin{figure}[h]
\centering
\includegraphics[width=0.41\textwidth]{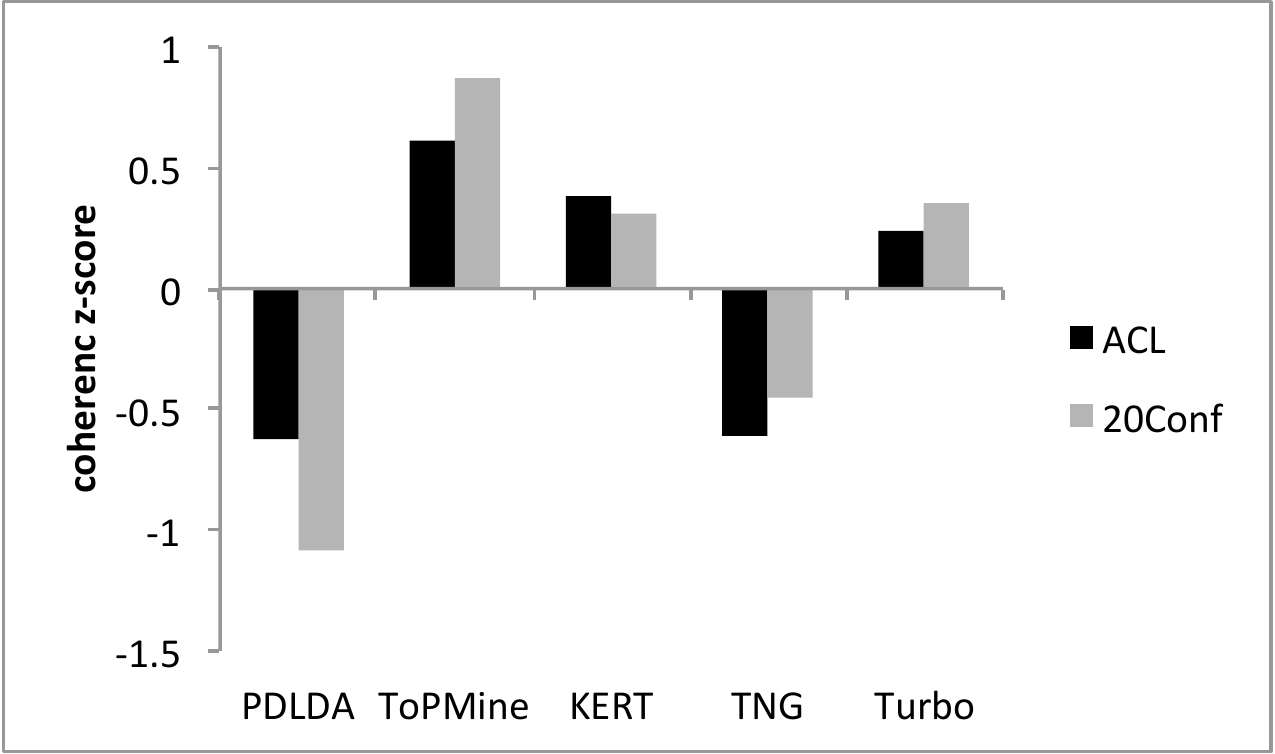}
\caption{Coherence of topics. Domain experts were asked to rate the `coherence' of each topic for each algorithm. Results were normalized into z-scores and averaged.}
\label{fig:coh}
\end{figure}

\begin{figure}[h]
\centering
\includegraphics[width=0.41\textwidth]{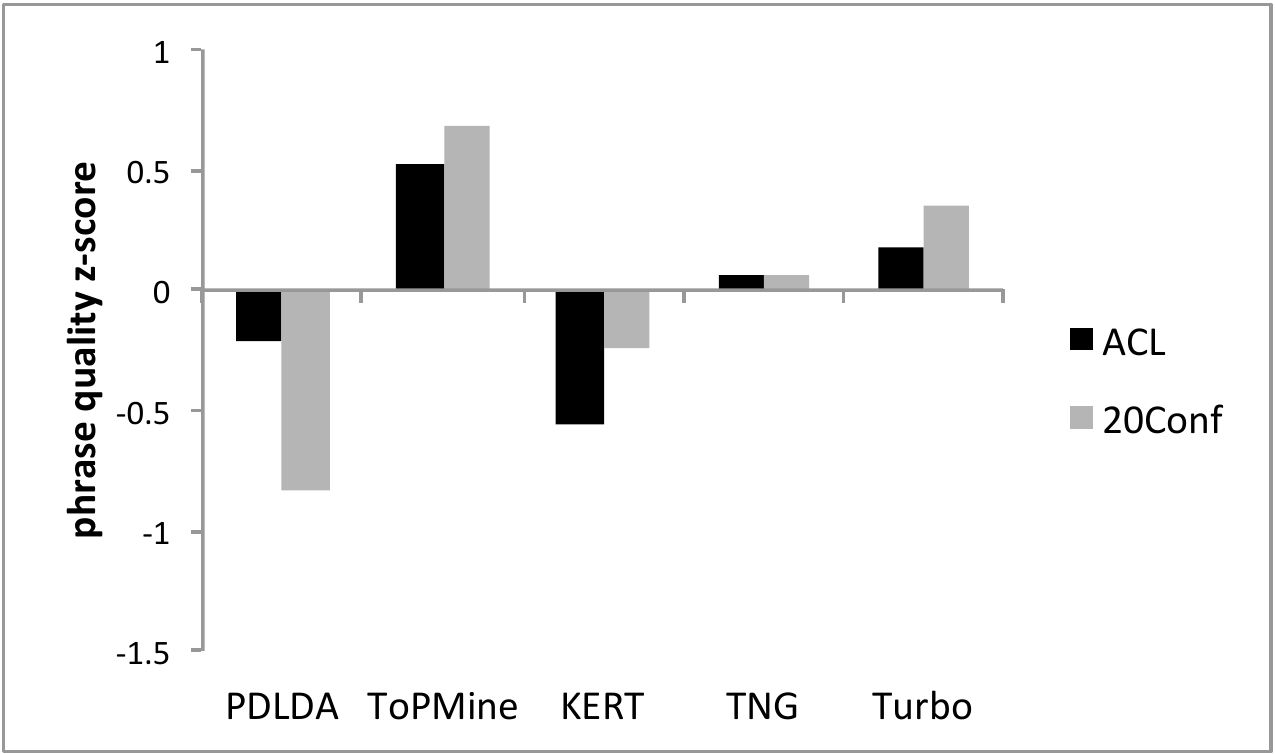}
\caption{Phrase quality. Domain experts were asked to rate the quality of phrases for each topic for each algorithm. Results were normalized into z-scores and averaged.}
\label{fig:pq}
\end{figure}

For each expert, ratings were standardized to a z-score. We compute each algorithm's topical scores by averaging that of five experts. The results are shown in Figure~\ref{fig:coh} and Figure~\ref{fig:pq}. 
\newpage
\subsubsection*{Discussion of Userstudy}
From Figures~\ref{fig:intrusion} and~\ref{fig:coh} we can tell that TopMine achieves similar performance to KERT in phrase intrusion, and demonstrates the best performance in topical coherence and phrase quality. We hypothesize that KERT's performance in phrase intrusion stems from its use of unconstrained frequent pattern mining and biased rankings towards longer phrases. Visual inspection suggests that many key topical unigrams are appended to common phrases, strengthening the notion of topical separation for all phrases. While this may aid KERT in phrase intrusion, we believe such practice lends to poor phrase quality, which is confirmed in Figure~\ref{fig:pq} as KERT demonstrates the lowest phrase-quality of the methods evaluated. A surprising occurrence is TNG and PD-LDA's poor performance in phrase intrusion. We suspect that this may be due to the many hyperparameters these complex models rely on and the difficulty in tuning them. In fact, the authors of PD-LDA make note that two of their parameters have no intuitive interpretation. Finally, Turbo Topics demonstrates above average performance on both datasets and user studies; this is likely a product of the rigorous permutation test the method employs to identify key topical phrases.

\subsection{Perplexity}
In addition to extracting meaningful and interpretable topical phrases, our ToPMine framework's PhraseLDA induces a statistical unigram topic model, upon the input corpus. To evaluate how well PhraseLDA's inference assumption that all words in our mined phrases should with high probability belong to the same topic, we evaluate how well the learned topic model predicts a held-out portion of our corpus. Because the generative process for PhraseLDA and LDA are the same, we can directly compare the perplexity between the two models to evaluate our method's performance.

\begin{figure}[h]
\centering
\includegraphics[width=0.47\textwidth]{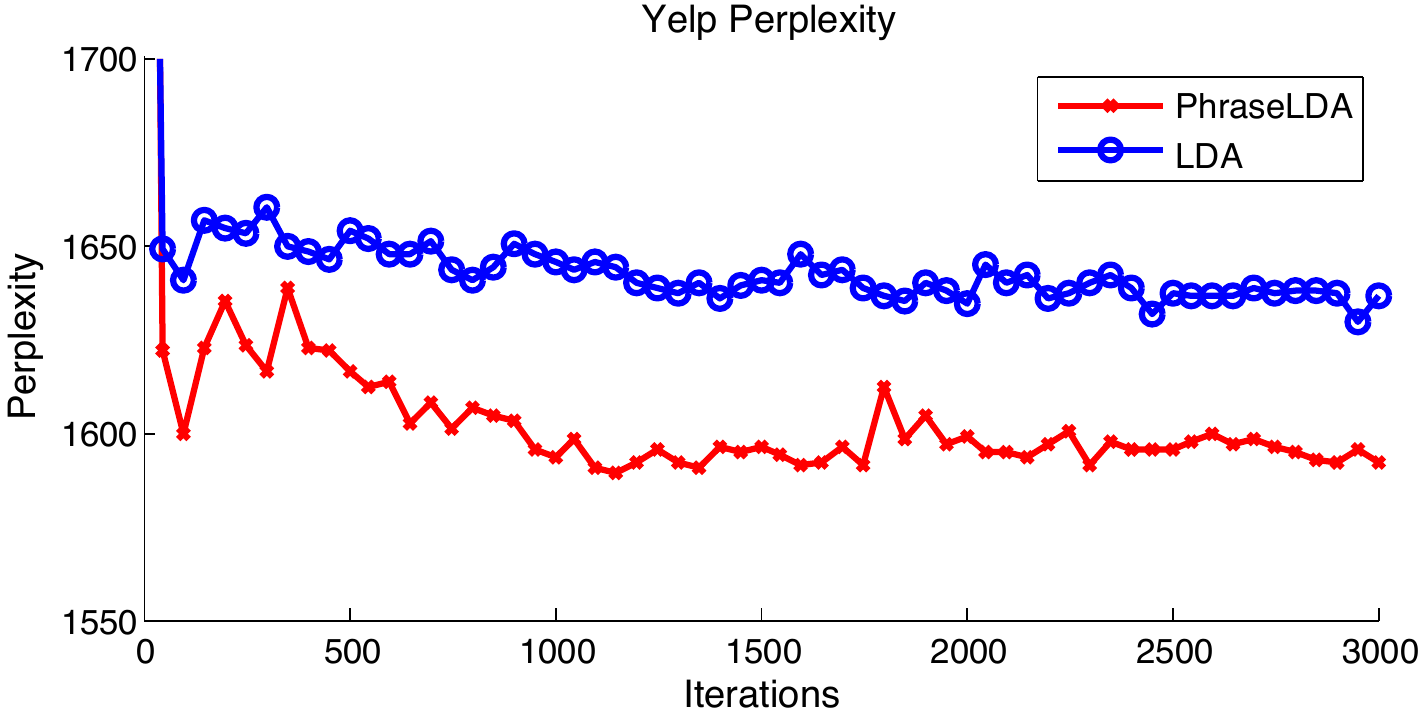}
\caption{Yelp Reviews. A comparison of the perplexity of LDA vs PhraseLDA during Gibbs sampling inference.}
\label{fig:Yelp}
\end{figure}

As we can see on the Yelp reviews dataset in Figure~\ref{fig:Yelp}, PhraseLDA performs significantly better than LDA demonstrating 45 bits lower perplexity than LDA. On the DBLP abstracts dataset, PhraseLDA demonstrates comparable perplexity to LDA. These results seem to validate the assumption that all words in our mined phrases should with high probability lie in the same topic. In addition, because our PhraseLDA can be seen as a more constrained version of LDA, these results provide an indication that our phrase mining method yields high-quality phrases as the perplexity of our learned model incorporating these phrases as constraints yields similar performance to LDA.

\begin{figure}[h]
\centering
\includegraphics[width=0.47\textwidth]{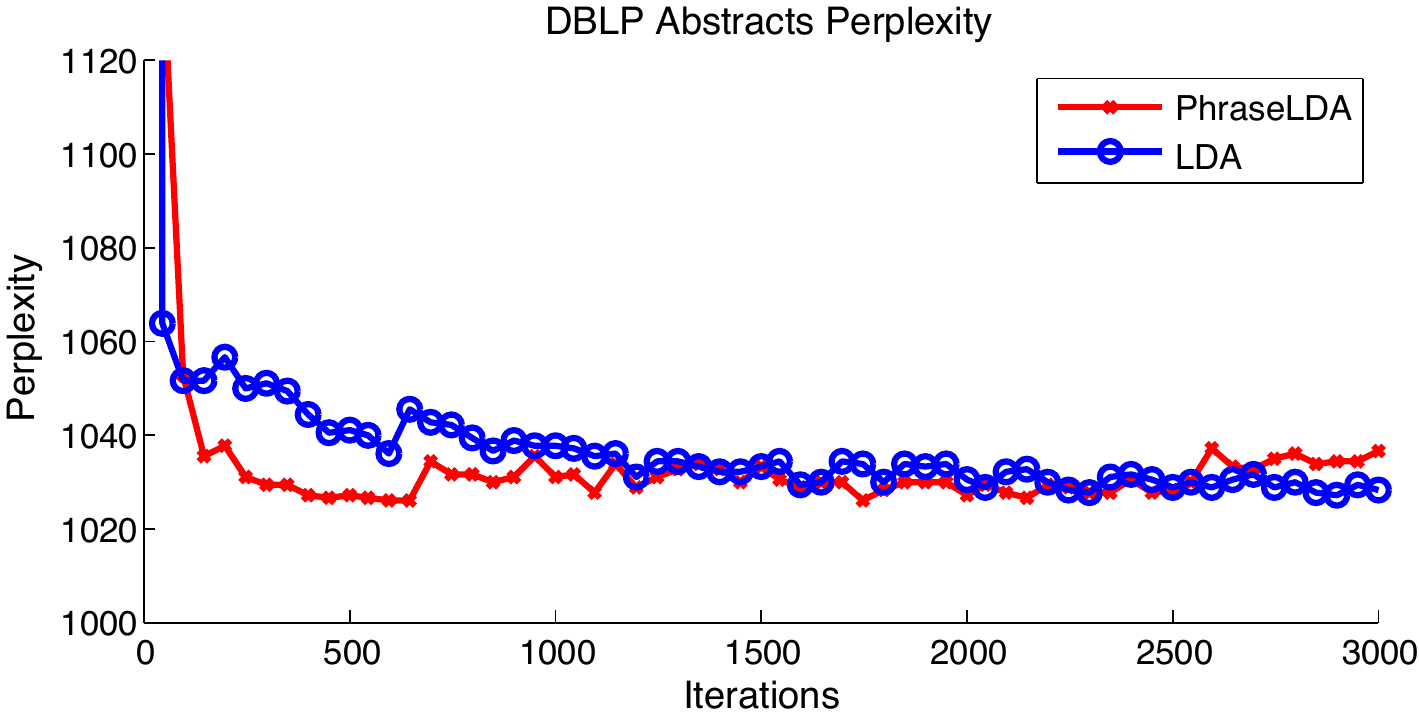}
\label{fig:abstract}
\caption{DBLP Abstracts. A comparison of the perplexity of LDA vs PhraseLDA during Gibbs sampling inference.}
\label{fig:perplexity}
\end{figure}
\subsection{Scalability}
\begin{figure}[h]
\centering
\includegraphics[width=0.41\textwidth]{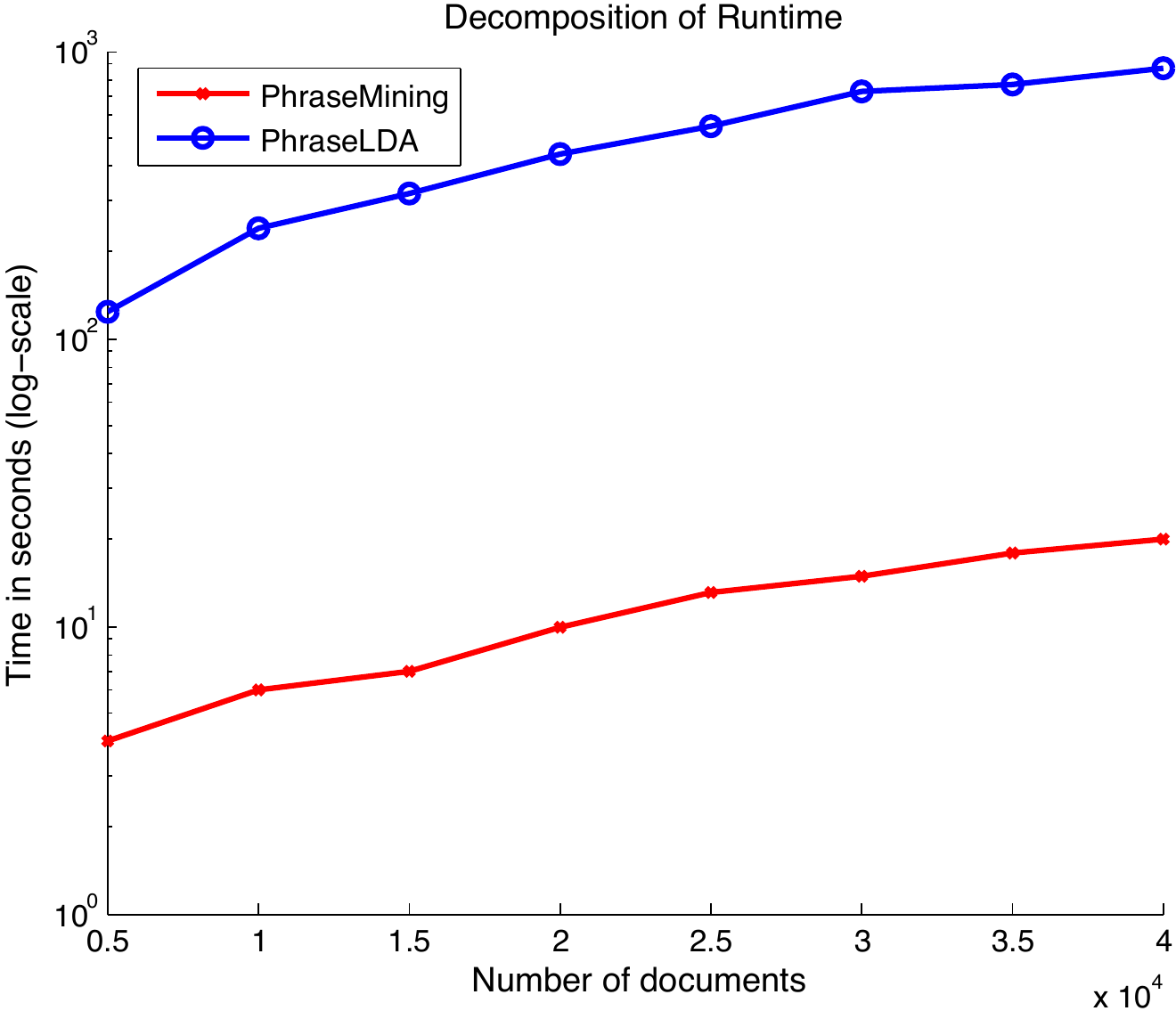}
\caption{Decomposition of our topical phrase mining algorithm into its two components: phrase mining and phrase-constrained topic modeling. The plot above, which is displayed on a log-scale, demonstrates the speed of the phrase-mining portion. With 10 topics and 2000 Gibbs sampling iterations, the runtime of the topic modeling portion is consistently 40X the phrase mining.}
\label{fig:decomp}
\end{figure}

To understand the run-time complexity of our framework, we first analyze the decomposition of ToPMine's runtime. On a high-level, ToPMine can be decomposed into two main separate procedures. The framework first involves frequent contiguous pattern mining followed by significant phrase construction. The second step is to take the `bag-of-phrases' output as constraints in PhraseLDA. By separately timing these two steps in our framework, we can empirically analyze the expected runtime of each step. Figure~\ref{fig:decomp} demonstrates the disparity in runtime between the phrase mining and topic modeling portions of ToPMine. Displayed on a log-scale for ease of interpretation we see that the runtime of our algorithm seems to scale linearly as we increase the number of documents (abstracts from our DBLP dataset). In addition, one can easily note that the phrase mining portion is of negligible runtime when compared to the topic modeling portion of the algorithm.

To evaluate our method's scalability to other methods, we compute our framework's runtime (on the same hardware) for datasets of various sizes and domains and compare them to runtimes of other state-of-the-art methods. For some datasets, competing methods could not be evaluated due to computational complexity leading to intractable runtimes or due to large memory requirements. We have attempted to estimate the runtime based on a smaller number of iterations whenever we face computational intractability of an algorithm on a specific dataset.
We used an optimized Java implementation MALLET\cite{mccallum2002mallet} for the TNG implementation and the topic modeling portions of KERT and Turbo Topics. For PD-LDA, we used the author's original C++ code. For LDA and PhraseLDA, the same JAVA implementation of PhraseLDA is used (as LDA is a special case of PhraseLDA). Because all these methods use Gibbs sampling to perform inference, we set the number of iterations to 1000. While we use hyperparameter optimization for our qualitative user-study tests and perplexity calculations, we do not perform hyperparameter optimization in our timed test to ensure a fair runtime evaluation. The runtime for ToPMine is the \textbf{full runtime} of the framework including both phrase mining and topic modeling.

Table~\ref{table:time} shows the runtime of each method on our datasets.  As expected, complex hierarchal models such as PD-LDA display intractable runtimes outside small datasets showing several magnitudes larger runtime than all methods except Turbo Topics. Turbo Topics displays a similar runtime due to the computationally intensive permutation tests on the back-off n-gram model. These methods were only able to run on the two sampled datasets and could not be applied to the full (larger) datasets. On short documents such as titles, KERT shows great scalability to large datasets barely adding any computational costs to LDA. Yet due to KERT's pattern-mining scheme, the memory constraints and the exponential number of patterns generated make large long-text datasets intractable. ToPMine is the only method capable of running on the full DBLP abstracts dataset with runtime in the same order as LDA. Under careful observation, PhraseLDA often runs in shorter time than LDA. This is because under our inference method, we sample a topic once for an entire multi-word phrase, while LDA samples a topic for each word. 

Tables~\ref{tab:DBLP},~\ref{tab:AP},~\ref{tab:YELP} are sample results of TopMine on three relatively large datasets - DBLP abstracts, AP News articles, and Yelp reviews. Our topical phrase framework was the only method capable on running on these three large, long-text datasets. In the visualization, we present the most probable unigrams from PhraseLDA as well as the most probable phrases below the unigrams. Automatic unstemming was performed as a post-processing step to visualize phrases in their most interpretable form. In many cases we see uninterpretable unigram topics that are made easier to interpret with the inclusion of topical phrases.  Overall we can see that for datasets that naturally form topics such as events in the news and computer science subareas, ToPMine yields high quality topical phrases. For noisier datasets such as Yelp, we find coherent, yet lower quality topical phrases. We believe this may be due to the plethora of background words and phrases such as `good', `love', and `great'. These and other words and phrases display sentiment and emphasis but are poor topical descriptors.
\begin{table}[h]
\fontsize{8pt}{8pt}\selectfont

\centering
\setlength{\extrarowheight}{2.0pt}
\begin{tabular}{ |L{0.9cm}| L{1.0cm}| L{1.458cm}| L{1.5cm}| L{1.57cm}|}
  \hline
 \emph{Method} & \emph{sampled dblp titles (k=5)} & \emph{dblp titles (k=30)} & \emph{sampled dblp abstracts} & \emph{dblp abstracts}\\
  \hline
  \centering
  PDLDA & 3.72(hrs) &$\sim$20.44(days) & 1.12(days) & $\sim$95.9(days)\\
  Turbo Topics & 6.68(hrs) & $>$30(days)* & $>$10(days)* & $>$50(days)*\\
  TNG & 146(s) & 5.57 (hrs) & 853(s) & NA\ding{61}\\
  LDA & \textbf{65(s)} & 3.04 (hrs) & 353(s) & 13.84(hours)\\
  KERT & 68(s) & 3.08(hrs) & 1215(s) & NA\ding{61}\\
  \textbf{ToPMine} & {67(s)} & \textbf{2.45(hrs)} & \textbf{340(s)} & \textbf{10.88(hrs)}\\
 \hline
\end{tabular}
\caption{We display the run-times of our algorithm on various datasets of different sizes from different domains. We sample $50$ thousand dblp titles and $20$ thousand dblp abstracts to provide datasets that the state-of-the art methods can perform on. For instances labeled \emph{*}, we estimate runtime by calculating the runtime for one topic and extrapolating for k topics. For instances labeled \emph{$\sim$} we extrapolate by calculating runtime for a tractable number of iterations and extrapolating across all iterations. For instances labeled \ding{61}, we could not apply the algorithm to the dataset because the algorithm exceeded memory constraints (greater than 40GB) during runtime.}
\label{table:time}
\end{table}

\begin{table*}[h!]
\fontsize{7pt}{9pt}\selectfont
\tabcolsep=0.10cm
\begin{tabular}{ @{} *6l @{} }
  \hline
 &\emph{Topic 1} & \emph{Topic 2} & \emph{Topic 3} & \emph{Topic 4} & \emph{Topic 5} \\
  \hline
 1-grams&problem & word & data& programming & data\\
 &algorithm & language &method &language & patterns\\
&optimal & text &algorithm & code & mining\\
&solution & speech &learning & type & rules\\
&search & system &clustering & object & set\\
&solve & recognition &classification & implementation & event\\
&constraints & character &based & system & time\\
&programming & translation &features & compiler & association\\
&heuristic & sentences &proposed& java & stream\\
&genetic & grammar &classifier& data & large\\
\hline
n-grams&genetic algorithm & natural language & data sets & programming language& data mining\\
&optimization problem & speech recognition &support vector machine & source code & data sets\\
&solve this problem & language model &learning algorithm & object oriented & data streams\\
&optimal solution & natural language processing &machine learning & type system & association rules\\
\newline
&evolutionary algorithm & machine translation &feature selection & data structure & data collection\\
&local search & recognition system &paper  we propose & program execution & time series\\
&search space & context free grammars &clustering algorithm & run time & data analysis\\
&optimization algorithm & sign language &decision tree & code generation & mining algorithms\\
&search algorithm & recognition rate &proposed method & object oriented programming & spatio temporal\\
&objective function & character recognition &training data & java programs & frequent itemsets\\
\end{tabular}
\caption{Five topics from a 50-topic run of ToPMine framework on our full DBLP abstracts dataset. Overall we see coherent topics and high-quality topical phrases we interpret as search/optimization, NLP, machine learning, programming languages, and data mining}
\label{tab:DBLP}
\end{table*}

\begin{table*}[h!]
\fontsize{7pt}{9pt}\selectfont
\centering
\tabcolsep=0.08cm
\begin{tabular}{@{} *6l @{} }
  \hline
 &\emph{Topic 1} & \emph{Topic 2} & \emph{Topic 3} & \emph{Topic 4} & \emph{Topic 5} \\
  \hline
1-grams&plant & church & palestinian & bush & drug\\
&nuclear & catholic & israeli & house & aid\\
&environmental & religious &israel &senate & health\\
&energy & bishop & arab& year & hospital\\
&year & pope & plo & bill & medical\\
&waste & roman &army & president & patients\\
&department & jewish &reported & congress & research\\
&power & rev &west & tax & test\\
&state & john &bank & budget & study\\
&chemical & christian &state & committee & disease\\
\hline
n-grams&energy department & roman catholic & gaza strip & president bush & health care\\
&environmental protection agency & pope john paul & west bank & white house & medical center\\
&nuclear weapons & john paul &palestine liberation organization & bush administration & united states\\
&acid rain & catholic church &united states & house and senate & aids virus\\
&nuclear power plant & anti semitism & arab reports & members of congress & drug abuse\\
&hazardous waste & baptist church &prime minister & defense secretary & food and drug administration \\
&savannah river & united states &yitzhak shamir & capital gains tax & aids patient\\
&rocky flats & lutheran church &israel radio & pay raise & centers for disease control\\
&nuclear power & episcopal church &occupied territories & house members & heart disease\\
&natural gas & church members &occupied west bank & committee chairman & drug testing\\
 \end{tabular}
\caption{Five topics from a 50-topic run of ToPMine on a large collection of AP News articles(1989). Overall we see high quality topical phrases and coherency of news topics such as environment, Christianity, Palestine/Israel conflict, Bush administration (senior), and health care}
\label{tab:AP}
\end{table*}
%

%
\begin{table*}[h!]
\fontsize{7pt}{9pt}\selectfont
\centering
\begin{tabular}{ @{} *6l @{} }
  \hline
 &\emph{Topic 1} & \emph{Topic 2} & \emph{Topic 3} & \emph{Topic 4} & \emph{Topic 5}\\
  \hline
1-grams&coffee & food & room & store & good\\
&ice & good &parking & shop & food\\
&cream & place &hotel & prices & place\\
&flavor & ordered &stay & find & burger\\
&egg & chicken & time &  place&ordered\\
&chocolate & roll &nice &buy & fries\\
&breakfast & sushi &place & selection & chicken\\
&tea & restaurant &great & items & tacos\\
&cake & dish &area & love & cheese\\
&sweet & rice &pool & great & time\\

 \hline
n-grams&ice cream & spring rolls & parking lot & grocery store & mexican food\\
&iced tea & food was good &front desk & great selection & chips and salsa\\
&french toast & fried rice &spring training & farmer's market & food was good\\
&hash browns & egg rolls &staying at the hotel & great prices & hot dog\\
&frozen yogurt & chinese food &dog park & parking lot & rice and beans\\
&eggs benedict & pad thai &room was clean & wal mart & sweet potato fries\\
&peanut butter & dim sum &pool area & shopping center & pretty good\\
&cup of coffee & thai food &great place & great place & carne asada\\
&iced coffee & pretty good &staff is friendly & prices are reasonable & mac and cheese\\
&scrambled eggs & lunch specials &free wifi & love this place & fish tacos\\
\end{tabular}

\caption{Five topics from a 10-topic run of our ToPMine framework on our full Yelp reviews dataset. Quality seems to be lower than the other datasets, yet one can still interpret the topics: breakfast/coffee, Asian/Chinese food, hotels, grocery stores, and Mexican food}
\label{tab:YELP}
\end{table*}

\
\section{Future Work}
\label{sec:future_work}
One natural extension to this work is to extend our topic model PhraseLDA to use a nonparametric prior over topics. This will systematically allow for a data-driven estimate of the number of underlying topics in the corpus. Another area of work is in further scalability of the topic model portion. Currently the decreased computational complexity stems from the efficient phrase-mining. By investigating other methods for topical inference, the overall time complexity of ToPMine may be significantly reduced. Another area of focus is to address how the minimum support criterion and pruning strategies treat similar phrases as separate discrete structures, counting them separately. While this phenomenon doesn't affect the `top-ranked' phrases, which have a count much larger than the minimum support, finding and merging similar phrases may lead to better recall and better topics. Further work may focus on strategies to identify and properly tie similar phrases. Finally, in Table~\ref{tab:DBLP} we notice background phrases like `paper we propose' and `proposed method' that occur in the topical representation due to their ubiquity in the corpus and should be filtered in a principled manner to enhance separation and coherence of topics.

\section{Conclusions}
\label{sec:conclusion}

In this paper, we presented a topical phrase mining framework, ToPMine, that discovers arbitrary length topical phrases. Our framework mainly consists of two parts: phrase mining and phrase-constrained topic modeling. In the first part, we use frequent phrase mining to efficiently collect necessary aggregate statistics for our significance score - the objective function that guides our bottom-up phrase construction. Upon termination, our phrase mining step segments each document into a bag of phrases. The induced partitions are incorporated as constraints in PhraseLDA allowing for a principled assignment of latent topics to phrases.

This separation of phrase-discovery from the topic model allows for less computational overhead than models that attempt to infer both phrases and topics and is a more principled approach than methods that construct phrases as a post-processing step to LDA. ToPMine demonstrates scalability on large datasets and interpretability in its extracted topical phrases beyond the current state-of-the-art methods.

\eat{
This separation of phrase-discovery from the topic model allows for less computational overhead than models that attempt to infer both phrases and topic. By combining our frequent phrase mining algorithm with a significance score-guided  bottom-up phrase construction algorithm, we can efficiently segment each document into a partition of phrases. Our topic model PhraseLDA efficiently utilizes the induced partitions to efficiently and systematically assign the same latent topic to each constituent word of a phrase. This framework has demonstrated scalability on large datasets and interpretability in its extracted topical phrases beyond the current state-of-the-art methods.
}

\afterpage{\clearpage}
\section{Acknowledgments}
This work was supported in part by the Army Research Lab under Cooperative Agreement No. W911NF-09-2-0053 (NSCTA), the Army Research Office under Cooperative Agreement No. W911NF-13-1-0193, National Science Foundation IIS-1017362, IIS-1320617, and IIS-1354329, HDTRA1-10-1-0120, and MIAS, a DHS-IDS Center for Multimodal Information Access and Synthesis at UIUC. Ahmed El-Kishky was sponsored by the National Science Foundation Graduate Research Fellowship grant NSF DGE-1144245.

\balance

%
\bibliographystyle{abbrv}
\bibliography{ToPMine}  
%
%

\section*{Appendix}
\small
\indent \indent In this section, we give the details of the collapsed gibbs sampling inference for PhraseLDA~\ref{sec:phraseLDA}. First, from equation~\ref{equ:collopsed}, we have
\begin{align*}
P(Z,W) &=  \frac{1}{C} \;P_{\text{\small{LDA}}}(Z,W) \prod_{d,g} f(\mathcal{C}_{d,g})\\
&\propto \prod_{k=1}^{K} \left(
\prod_{d=1}^{D}\Gamma(\alpha_k+ \mathcal{N}_{d,k}) \frac{\prod_{x=1}^{V} \Gamma(\beta_x+ \mathcal{N}_{x,k})}
{\Gamma(\sum_{x=1}^{V} \beta_x+ \mathcal{N}_{k})}
\right)
\end{align*}
\noindent where the derivation of second line can be found in~\cite{griffiths2002gibbs}.

Second,
\begin{align*}
 &p(\mathcal{C}_{d,g} = k|W,Z_{\setminus {\mathcal{C}_{d,g}}})\; \propto \; p(Z,W) \;
  \propto \;   \frac{ \Gamma(\alpha_{k}+\mathcal{N}_{d,k\setminus \mathcal{C}_{d,g}} + W_{d,g})}
 { \Gamma(\alpha_{k}+\mathcal{N}_{d,k\setminus \mathcal{C}_{d,g}}) }*  \\
&\prod_{j=1}^{W_{d,g}}
\frac{\Gamma(\beta_{w_{d,g,j}}+\mathcal{N}_{w_{d,g,j},k\setminus \mathcal{C}_{d,g}}+1 )}
{\Gamma(\sum_{x=1}^{V} \beta_x+\mathcal{N}_{k\setminus \mathcal{C}_{d,g}}+W_{d,g})}
/
\frac{\Gamma(\beta_{w_{d,g,j}} + \mathcal{N}_{w_{d,g,j},k\setminus \mathcal{C}_{d,g}} )}
{\Gamma (\sum_{x=1}^{V} \beta_x + \mathcal{N}_{k\setminus \mathcal{C}_{d,g}} )}
 \nonumber \\
&=\prod_{j=1}^{W_{d,g}}
(\alpha_{k} + \mathcal{N}_{d,k\setminus \mathcal{C}_{d,g}}+ j - 1)
\frac{
( \beta_{w_{d,g,j}} + \mathcal{N}_{k,w_{d,g,j}\setminus \mathcal{C}_{d,g}})
}{
(\sum_{x=1}^{V} \beta_x +\mathcal{N}_{k\setminus \mathcal{C}_{d,g}} + j - 1)
}
\end{align*}
where we utilize the fact that $\Gamma(x+1)=x\Gamma(x)$.
\ \\

\end{document}